\documentclass[letterpaper]{article} 
\usepackage{aaai23}  
\usepackage{times}  
\usepackage{helvet}  
\usepackage{courier}  
\usepackage[hyphens]{url}  
\usepackage{graphicx} 
\usepackage{natbib}  
\usepackage{caption} 
\usepackage{algorithm}
\usepackage{algorithmic}

\usepackage{newfloat}
\usepackage{listings}
\DeclareCaptionStyle{ruled}{labelfont=normalfont,labelsep=colon,strut=off} 
\floatstyle{ruled}
\newfloat{listing}{tb}{lst}{}
\floatname{listing}{Listing}
\usepackage{amsmath}
\usepackage{amssymb}
\usepackage{multirow}
\usepackage{booktabs}
\title{PaRot: \textbf{Pa}tch-Wise \textbf{Rot}ation-Invariant Network via Feature Disentanglement and Pose Restoration}
\author{
    Dingxin Zhang\equalcontrib,
    Jianhui Yu\equalcontrib,
    Chaoyi Zhang,
    Weidong Cai
}
\affiliations{
    School of Computer Science, University of Sydney, Australia

    dzha2344@uni.sydney.edu.au, \{jianhui.yu, chaoyi.zhang, tom.cai\}@sydney.edu.au
}

\begin{document}

\maketitle

\begin{abstract}
Recent interest in point cloud analysis has led rapid progress in designing deep learning methods for 3D models.
However, state-of-the-art models are not robust to rotations, which remains an unknown prior to real applications and harms the model performance.
In this work, we introduce a novel \textbf{Pa}tch-wise \textbf{Rot}ation-invariant network (PaRot), which achieves rotation invariance via feature disentanglement and produces consistent predictions for samples with arbitrary rotations.
Specifically, we design a siamese training module which disentangles rotation invariance and equivariance from patches defined over different scales, e.g., the local geometry and global shape, via a pair of rotations.
However, our disentangled invariant feature loses the intrinsic pose information of each patch.
To solve this problem, we propose a rotation-invariant geometric relation to restore the relative pose with equivariant information for patches defined over different scales.
Utilising the pose information, we propose a hierarchical module which implements intra-scale and inter-scale feature aggregation for 3D shape learning.
Moreover, we introduce a pose-aware feature propagation process with the rotation-invariant relative pose information embedded.
Experiments show that our disentanglement module extracts high-quality rotation-robust features and the proposed lightweight model achieves competitive results in rotated 3D object classification and part segmentation tasks.
Our project page is released at: https://patchrot.github.io/.
\end{abstract}

\section{Introduction}

Point cloud analysis has recently drawn much interest from researchers. 
As a common form of 3D representations, point clouds are applied in areas such as unmanned driving and 3D face recognition. 
Recent deep learning models~\cite{PointNet, PointNet++} show great potential on well aligned point clouds for classification and segmentation. 
However, 3D objects are normally rotated and orientation angles are unknown in real scenarios, which can largely impact the deep learning models that are sensitive to rotations. 
Therefore, making the model invariant to rotations becomes an important research topic.

\begin{figure}[t]
\begin{center}
  \includegraphics[width=1.0\linewidth]{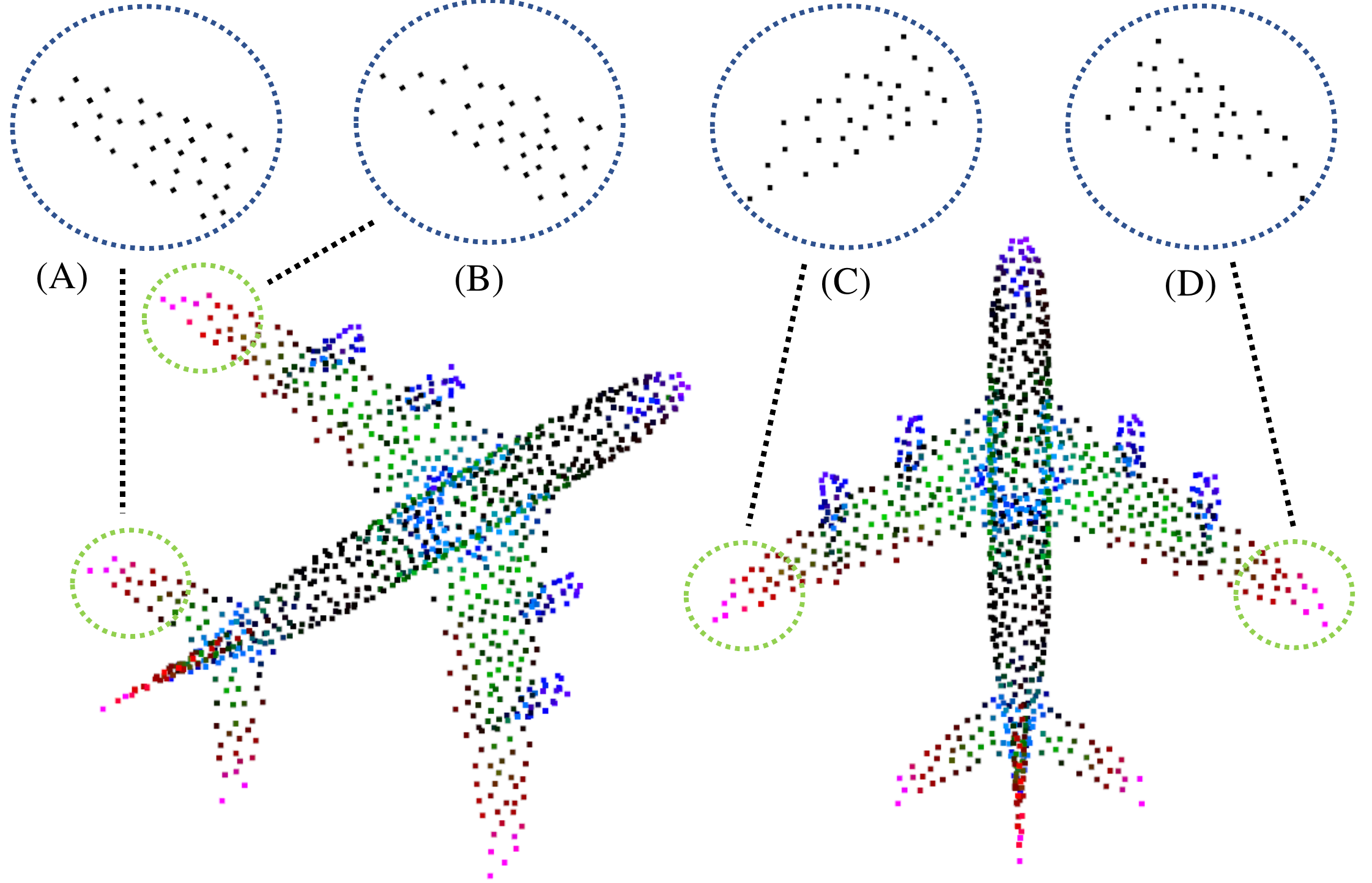}
\end{center}
  \caption{Illustration of pose information loss when generating patch-wise rotation-invariant features. 
  We visualise our learned patch-wise rotation-invariant features of an air plane instance under two different orientations. 
  As it can be seen, for B and C that have exactly same geometric shapes and different poses (i.e. orientation and global position), the same features are generated. 
  However, the features for patches on tail (A), left wing (B, C) and right wing (D) will be very similar. 
  The information for distinguishing their difference between are lost.
  }
\label{fig:pose}
\end{figure}

Pioneering work has attempted to obtain rotation robustness by transforming the shape into a canonical pose~\cite{PointNet, stn}, which cannot achieve consistent invariance to rotation.
Recent works construct rotation-invariant representations from local geometry as model input~\cite{RIConv, RIgcn}, which achieves consistent behavior under random rotations.
However, the rotation-invariant representations lose intrinsic pose information (i.e., orientation and position) as illustrated in Fig.~\ref{fig:pose}.
To solve this problem, \citet{RIF} and \citet{localmeetglobal} generate rotation-invariant features from the global 3D shape and restore the global pose information by simply concatenating features at both local and global scales, without fully exploring how the pose information can be used more effectively.
Moreover, these methods achieve rotation invariance based on handcrafted features invariant to rotation, which could limit the model performance.

In this work, we borrow the idea of \citet{localmeetglobal} where we obtain rotation-invariant information from both local and global scales.
Specifically, we capture point patches from the local geometry and global shape.
We then apply a feature disentanglement module, where pairs of rotations are introduced to each point patch, to extract rotation-invariant shape content and rotation-equivariant orientation information via a siamese training process.
In this way, the invariant features are dynamically generated, which enhances the feature representation.
To recover the pose information, we firstly define a geometric relation between two patches by computing the positional and orientational differences.
We then propose a \textbf{Pa}tch-wise \textbf{Rot}ation-invariant network (PaRot), which takes as input the geometric relation and rotation-invariant shape contents encoded from local patches, for an intra-scale aggregation process.
In addition, an inter-scale process is considered, which takes the geometric relation and features encoded across local and global patches, for global context exploiting.
Moreover, we follow PointNet++~\cite{PointNet++} for segmentation by using and modifying the feature propagation module at a minimal cost.
Specifically, we propose a pose-aware feature propagation module, where the previously static distance-based feature interpolation is replaced by a learnable process, with encoded geometric relations to preserve the rotation-invariant relative pose information.
More importantly, extensive experiments on different benchmarks present the superiority of our method.

The contributions of this work are summarised as follows: 
(1) We propose a siamese training module by introducing pairs of rotations to disentangle patch-wise learnable high-quality rotation-invariant shape content feature and rotation-equivariant orientation feature;
(2) We define a rotation-invariant geometric relation representation to restore relative pose information between patches to guide the inter-scale and intra-scale learning;
(3) We design a relative pose-aware feature propagation method for more accurate rotation-invariant segmentation.

\section{Related Work}

\subsubsection{Deep Learning on Point Clouds.}
Previous deep learning methods for 3D point clouds capitalise on the advanced development of 2D convolutional neural networks.
Recent works directly consume point set data by extracting point-wise feature. 
PointNet~\cite{PointNet} presents a groundbreaking structure, utilising MLPs to learn point-wise spatial features and achieve permutation invariance with max pooling. The following works are extended on the basis of PointNet framework, including learning local context to abstracting geometry information from different scales of patches~\cite{PointNet++, pointweb, medical}, developing convolution operators for better feature extraction~\cite{KPconv, RSCNN}, and improving symmetry functions to promote feature aggregation~\cite{curvenet, recyclemax}. However, most methods are rotation-sensitive and their performances degrade drastically when input point clouds are rotated arbitrarily.

\subsubsection{Rotation Equivariance.} 
One approach to achieving rotation robustness is to ensure the learned features of all intermediate layers rotate correspondingly with the input.
Spherical convolution-based methods~\cite{spfcnn, Spherical, SFCNN} transform point clouds into a spherical harmonic domain and apply spherical convolutions to capture roughly rotation-equivariant features. 
Tensor field-based networks~\cite{TFN2, Quaternion, VN} consume and output tensor field features that maintain strictly rotation-equivariant.
To ensure rotation invariance, these methods require an extra operation to transform high-level equivariant features into an invariant form, which will introduce information loss during training.
In our work, the equivariant orientation features are employed for restoring relative pose information during hierarchical geometric learning to reduce the information loss.


\subsubsection{Rotation Invariance.}
Another approach focuses on learning rotation-invariant features.
A common approach is to transform the Cartesian coordinates of point clouds into a handcrafted rotation-invariant representation in the data pre-processing stage.
\citet{RIConv}, \citet{ClusterNet}, and \citet{SGMNet} design handcrafted features within local patches.
These methods eliminate pose information of patches when generating rotation-invariant features and harm the geometry learning process.
To address this issue, \citet{Globalcontext}, \citet{RIF}, and \citet{Pari} take relative pose information into consideration when handcrafting representations. 
The relative pose between neighbouring patches~\cite{Pari} or between local patches and global shapes~\cite{Globalcontext, RIF, localmeetglobal} is embedded into handcrafted features.
In our work, the patch-wise rotation-invariant features are abstracted via neural networks and arbitrary rotations. Meanwhile, pose information is preserved by predicting patch-wise orientations and restored by computing intra- and inter-scale geometric relations.
Moreover, we embed geometric relations in feature propagation process to enhance the segmentation performance.

\subsubsection{Siamese Training.}
Siamese training enables feature disentanglement, which is an efficient technique in exploring the data variation and similarity.
Recent works employ siamese training in disentangling rotation-equivariant and rotation-invariant features~\cite{canonicalcapsule, wildsiamese, detarnet, condor}.
However, those methods mainly focus on registration and reconstruction tasks, and the extracted equivariant features are used for canonicalization. In contrast, equivariant orientation matrices in our method are employed to construct geometric relations for relative pose restoration, so that we can aggregate rotation-invariant features from intra- and inter-scale learning.




\section{Method}
\begin{figure*}[t]
\centering
\includegraphics[width=1.0\linewidth]{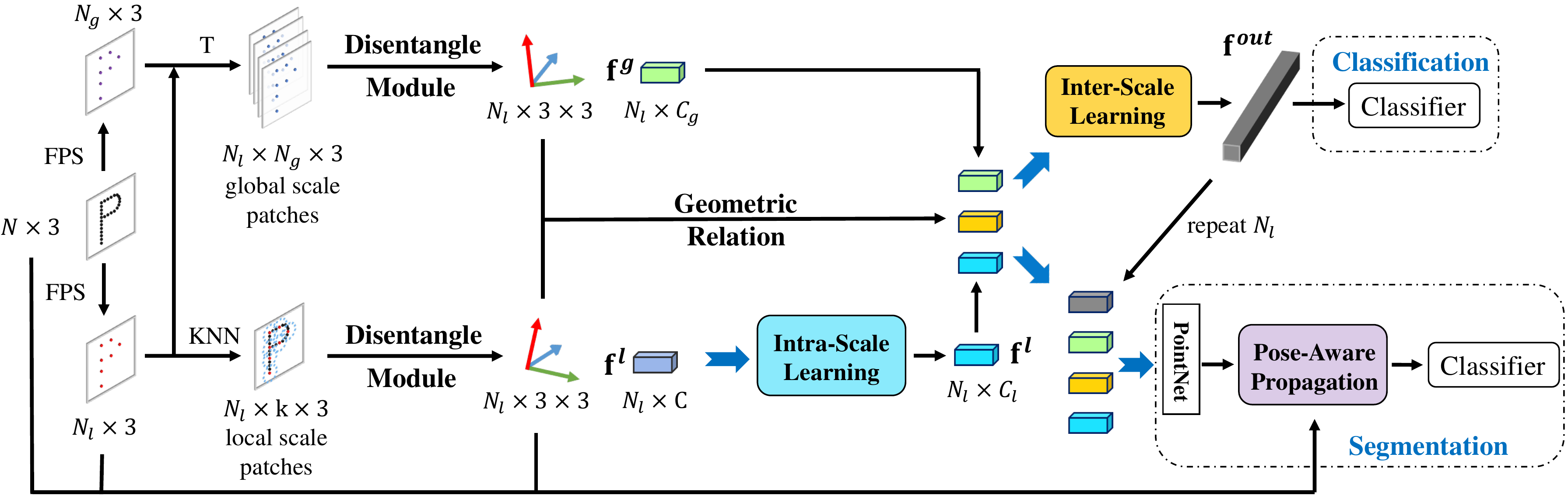}
\caption{The overall architecture of PaRot for 3D classification and segmentation, where KNN and T denote $k$ nearest neighboring and translation, respectively.
We generate local scale patches and global scale patches with FPS, KNN, and translation operations. Then we disentangle patches of different scales into shape content information and orientation information separately. The learned orientations are utilised to determine geometric relations for restoring patch-wise relative pose and guiding intra- and inter-scale invariant features learning and pose-aware feature propagation.
}
\label{fig:short}
\end{figure*}
Given a point cloud $\mathbf{P} \in \mathbb{R}^{N \times 3}$, a rotation-robust point cloud model $f(\cdot)$ needs to be invariant to any arbitrary rotation $\mathbf{R} \in \mathbb{R}^{3 \times 3}$ applied to $\mathbf{P}$ and produces consistent predictions: $f(\mathbf{P}) = f(\mathbf{P}\mathbf{R})$.
A Patch-wise Rotation-invariant network (PaRot) is introduced to achieve this goal. We first disentangle patch-wise rotation invariance and equivariance from shape descriptors (Section~\ref{sec:3.1}).
Comprehensive geometric representations with rotation invariance are extracted intra-scale and inter-scale, with geometric relations embedded to preserve pose relations between different patches (Section~\ref{sec:3.2}).
Finally, we propose a rotation-invariant feature propagation module with geometric relations to maintain the rotation invariance for semantic point labelling (Section~\ref{sec:3.3}).
\subsection{Rotation Invariance and Equivariance Disentanglement} \label{sec:3.1}
\begin{figure}[t]
    \centering
   \includegraphics[width=\linewidth]{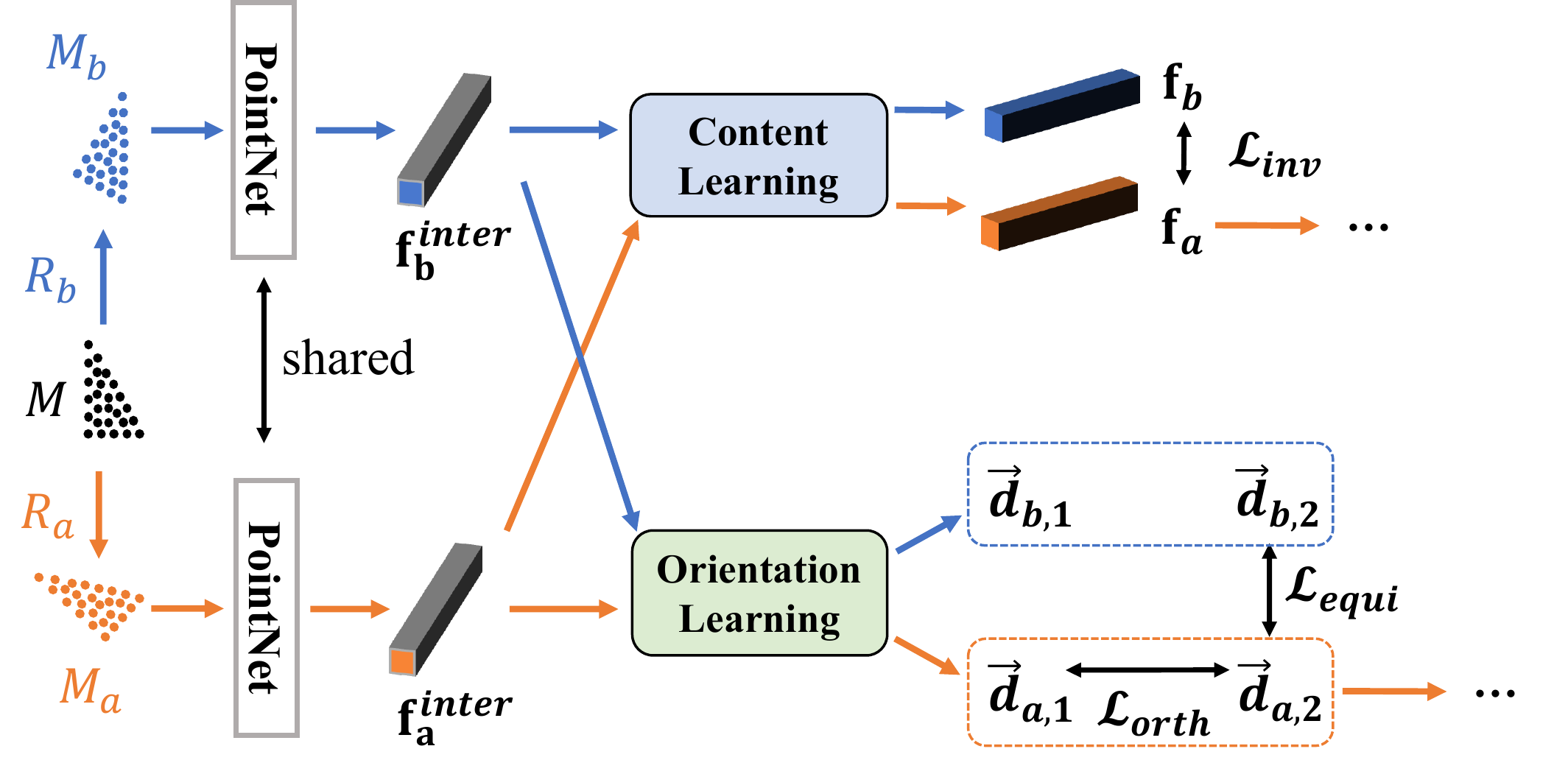}
   \caption{Frameworks of disentanglement module. The input patch is randomly rotated twice and then are independently fed through into the module as branch $a$ and $b$.
   The output of branch $a$ is hierarchically encoded while that of branch $b$ is used to assist learning.}
\label{fig:disentangle}
\end{figure}
Inspired by \cite{canonicalcapsule}, we propose a disentanglement module based on the siamese training pipeline, which decomposes latent shape descriptors into rotation-invariant shape contents for rotation invariance study and rotation-equivariant shape orientations to preserve pose information.

\subsubsection{Rotation-Invariant Content Learning.}
We discuss that for any 3D object under rotations, the shape content remains invariant to random rotations, and we extract such rotation-invariant content features in a patch-wise manner.
Particularly, as shown in Fig.~\ref{fig:disentangle}, we introduce a pair of arbitrary rotations $\mathbf{R}_a$ and $\mathbf{R}_b$ to input point patch $\mathbf{M} \in \mathbb{R}^{n \times 3}$, leading to two randomly rotated patches $\mathbf{M}_{a}$ and $\mathbf{M}_{b}$.
A light-weight PointNet network~\cite{PointNet} is employed and shared between $\mathbf{M}_{a}$ and $\mathbf{M}_{b}$ to encode the geometric information, leading to two intermediate shape descriptors $\mathbf{f}_{a}^{inter}$ and $\mathbf{f}_{b}^{inter}$.
Multi-layer perceptrons (MLPs) are thus applied to disentangle rotation-invariant shape contents $\mathbf{f}_{a}$ and $\mathbf{f}_{b}$ from latent shape descriptors, and rotation invariance is achieved via minimizing the feature distance between $\mathbf{f}_{a}$ and $\mathbf{f}_{b}$.
Hence, we define a \textit{rotation-invariant} loss function $\mathcal{L}_{inv}$ to enforce a high degree of similarity between these two features under rotations, which is represented as:
\begin{equation} \label{eq:inv_loss}
\mathcal{L}_{inv} = \left \| \mathbf{f}_{a} - \mathbf{f}_{b} \right\|_2^2,
\end{equation}
where $\left \| \cdot \right\|_2^2$ denotes the L2 Norm.

\begin{figure}[t]
\begin{center}
   \includegraphics[width=1.0\linewidth]{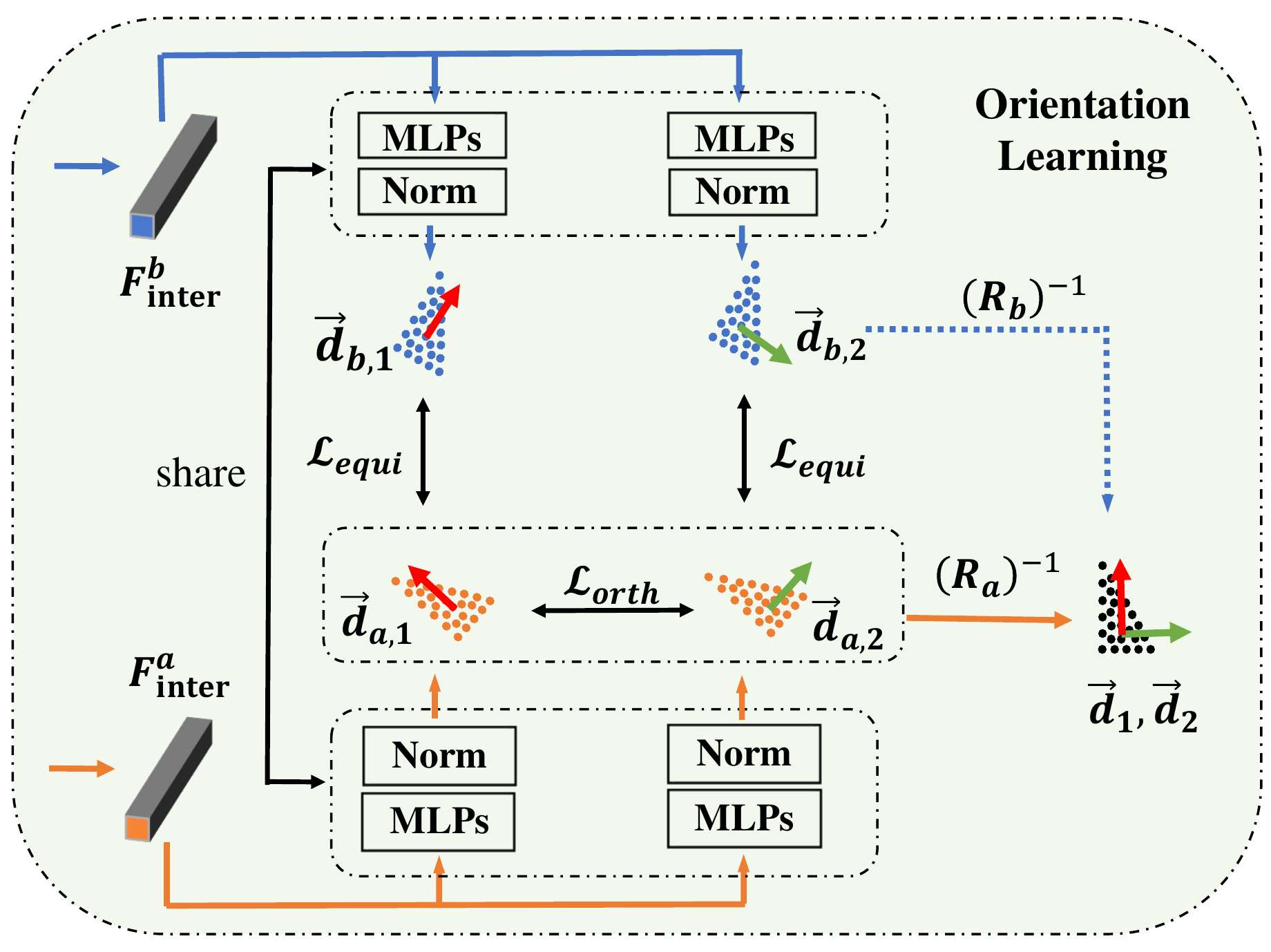}
\end{center}
   \caption{Illustrations of rotation-equivariant orientation learning. The output orientation of branch $a$ is rotated back by $\mathbf{R}_{a}^{-1}$ which is used as the predicted orientation of the original patch $\mathbf{M}$.}
\label{fig:orientation}
\end{figure}

\subsubsection{Rotation-Equivariant Orientation Learning.}
An obvious intuition we discuss here is that for any 3D patch, there exists an intrinsic orientation, essential for recovering the pose and denoted by a rotation matrix, which is naturally equivariant to rotations.
To this end, we borrow the idea from FS-Net~\cite{FSnet} to extract direction vectors from latent shape descriptors.
Specifically, as shown in Fig.~\ref{fig:orientation}, two perpendicular direction vectors $\vec{\mathbf{d}}_{1}$ and $\vec{\mathbf{d}}_{2}$ are predicted for each branch in terms of the latent shape descriptors $\mathbf{F}$, where we drop subscripts (i.e., $a$ and $b$) indexing the siamese branch when same operations are applied to both branches.
To ensure the equivariance, we introduce a \textit{rotation-equivariant} loss function $\mathcal{L}_{equi}$ between the learned direction vectors as follows:
\begin{equation} \label{eq:equi_loss}
\small
    \mathcal{L}_{equi} = \left \| \vec{\mathbf{d}}_{a, 1} - \mathbf{R}_{b}^{-1}\mathbf{R}_{a} \vec{\mathbf{d}}_{b, 1} \right\|_{2}^{2} + \left \| \vec{\mathbf{d}}_{a, 2} - \mathbf{R}_{b}^{-1}\mathbf{R}_{a}\vec{\mathbf{d}}_{b, 2} \right\|_{2}^{2},
\end{equation}
where $\mathbf{R}_{b}^{-1}\mathbf{R}_{a}$ is a rotation matrix transforming $\mathbf{M}_{a}$ to $\mathbf{M}_{b}$.
In this way, the direction vectors are enforced to contain only orientation information of the local patch, which are later used for pose information embedding.
Furthermore, to preserve the orthogonality between $\vec{\mathbf{d}}_{1}$ and $\vec{\mathbf{d}}_{2}$ as the column vectors of a rotation matrix, we design a loss function $\mathcal{L}_{orth}$ represented as follows:
\begin{equation} \label{eq:orth_loss}
\mathcal{L}_{orth} = \left \|{\vec{\mathbf{d}}_{1}}^{\top} \vec{\mathbf{d}}_{2} \right\|_{2}^{2}.
\end{equation}

One thing needs to be mentioned is that our orientation matrix is learned based on the random rotated pose of $\mathbf{M}$, which is different from that of the initial input $\mathbf{M}$.
As $\mathbf{R}_{a}$ and $\mathbf{R}_{b}$ are only introduced for disentanglement, we need to remove their impacts and obtain the initial orientation $\mathbf{O}$ of $\mathbf{M}$ by transforming predicted direction vectors back as: $[\vec{\mathbf{d}}_{1}, \vec{\mathbf{d}}_{2}] = \mathbf{R}_{a}^{-1}[\vec{\mathbf{d}}_{a, 1}, \vec{\mathbf{d}}_{a, 2}] = \mathbf{R}_{b}^{-1} [\vec{\mathbf{d}}_{b, 1}, \vec{\mathbf{d}}_{b, 2}$].

As $\vec{\mathbf{d}}_{1}$ and $\vec{\mathbf{d}}_{2}$ are ensured to be non-parallel, we can simply define a third direction vector $\vec{\mathbf{d}}_{3} = \vec{\mathbf{d}}_{1} \times \vec{\mathbf{d}}_{2}$ which is orthogonal to both $\vec{\mathbf{d}}_{1}$ and $\vec{\mathbf{d}}_{2}$.
We hence denote the orientation matrix learned for initial input $\mathbf{O}$ as the concatenation of direction vectors: $\mathbf{O} = [\vec{\mathbf{d}}_{1}, \vec{\mathbf{d}}_{2}, \vec{\mathbf{d}}_{3}]$, where $[\cdot]$ is concatenation.
Finally, as all learnable MLPs are shared across both branches, our learned rotation-invariant shape contents and orientations are the same.
Following the conventional implementation of the siamese training procedure~\cite{canonicalcapsule}, we utilise the outputs of branch $a$ for the later model inference and only calculate $\mathcal{L}_{orth}$ for branch $a$.



\subsection{Intra- and Inter-Scale Rotation Invariance Learning with Geometric Relations} \label{sec:3.2}
For rotation invariance learning, we propose a geometric relation embedding network to aggregate rotation-invariant features from intra- and inter-scale learning, where geometric relations between patches are computed with patch-wise orientation matrices and global absolute positions to recover relative pose information.

\subsubsection{Geometric Relation Representation.}
\begin{figure}[t]
\begin{center}
   \includegraphics[width=0.9\linewidth]{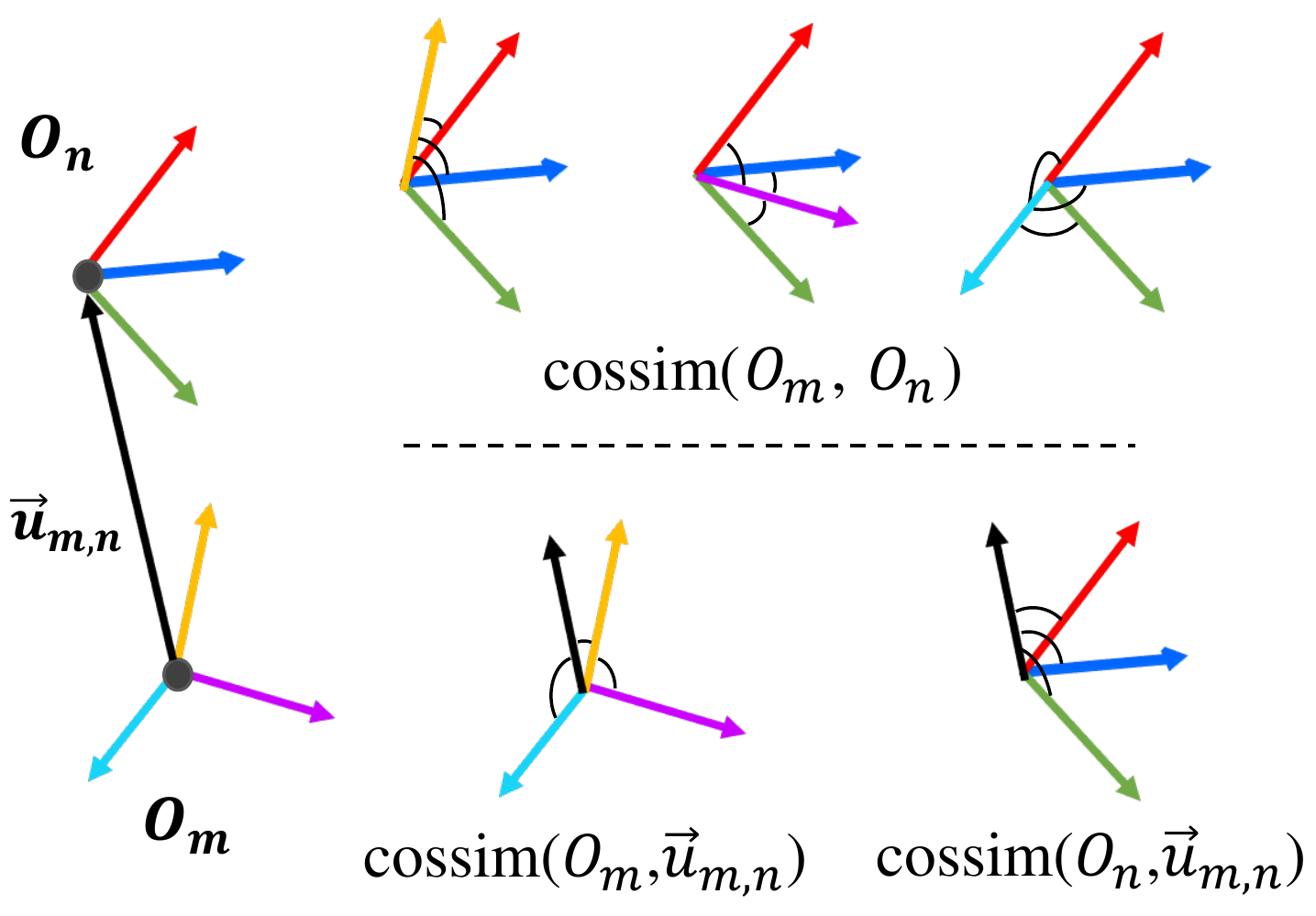}
\end{center}
   \caption{Angles used in geometric relation representation. 
   For two patches $\mathbf{M}_m$ and $\mathbf{M}_n$, we use different colors to illustrate different vectors in their orientation matrices $\mathbf{O}_{m}$ and $\mathbf{O}_{n}$.}
\label{fig:representation}
\end{figure}
The pose information of patches is critical for geometric feature learning, but it cannot be encoded into a rotation-invariant form, since the orientations and global positions are equivariant to rotations.
To maintain such pose information, we have to consider geometric properties between different patches that are invariant to rotations, such that the learned rotation invariance can be maintained.
To achieve this goal, we adopt relative geometric relations, i.e., relative angles and distances between patches, to retain the pose information of the original 3D shape during feature learning.
Specifically, given two patches $\mathbf{M}_m$ and $\mathbf{M}_n$ with corresponding reference points $\mathbf{p}_m$ and $\mathbf{p}_n$, their orientation matrices $\mathbf{O}_{m}=[\vec{\mathbf{d}}_{1}^{m}, \vec{\mathbf{d}}_{2}^{m}, \vec{\mathbf{d}}_{3}^{m}]$ and $\mathbf{O}_{n}=[\vec{\mathbf{d}}_{1}^{n}, \vec{\mathbf{d}}_{2}^{n}, \vec{\mathbf{d}}_{3}^{n}]$ can be learned according to our disentanglement module.
Hence, the geometric relation $\mathcal{G}(\mathbf{M}_m, \mathbf{M}_n)$ between the two patches is defined as:
\begin{equation} \label{eq:geo_relation}
\begin{aligned}
\mathcal{G}(\mathbf{M}_m, \mathbf{M}_n) =& [\operatorname{dist}(\mathbf{p}_m, \mathbf{p}_n), \operatorname{cossim}(\mathbf{O}_{m}, \mathbf{O}_{n}),\\ 
&\operatorname{cossim}(\mathbf{O}_{m}, [\vec{\mathbf{u}}_{mn}]),\\
&\operatorname{cossim}(\mathbf{O}_{n}, [\vec{\mathbf{u}}_{mn}])],
\end{aligned}
\end{equation}
where $\operatorname{dist}(\cdot)$ calculates the Euclidean distance between two points, $\operatorname{cossim}(\cdot)$ computes the cosine similarity between all column vectors of first matrix and all columns vectors of second matrix and $\vec{\mathbf{u}}_{mn} = \mathbf{p}_n - \mathbf{p}_m$ is the vector between the reference points of two patches, pointing from $\mathbf{p}_n$ to $\mathbf{p}_m$.
As shown in Fig.~\ref{fig:representation}, $\mathcal{G}(\mathbf{M}_m, \mathbf{M}_n)$ consists of 1 distance parameter, and $9+3+3=15$ angles parameters.

\subsubsection{Local-Scale Feature Disentangling.}
To generate local patches, we sample $N_{\ell}$ points using farthest point sampling from the original point set $\mathbf{P}$ as a local query point set $\mathbf{Q} \in \mathbb{R}^{N_{\ell} \times 3}$. 
For any point $\mathbf{q}_{i} \in \mathbf{Q}$ taken as the reference point, $k$-NN search is performed on $\mathbf{P}$, resulting in a total of $N_{\ell}$ patches.
Then, we generate a local reference frame for each patch and implement patch-wise translations to ensure the reference point $\mathbf{q}_{i}$ of patch $\mathbf{M}_{i}^{\ell}$ coincide with the reference frame origin point:
\begin{equation} \label{translate}
\begin{aligned}
\mathbf{M}_{i}^{\ell} = [q_j - q_i]_{j: j \in \mathcal{N}_{\mathbf{P}}(i)},
\end{aligned}
\end{equation}
where $\mathcal{N}_{\mathbf{P}}(i)$ denotes the neighboring points of $\mathbf{q}_{i}$ in $\mathbf{P}$.
We then apply shape feature disentanglement to extract rotation-invariant content features $\mathbf{f}_{i}$ and rotation-equivariant orientation matrices $\mathbf{O}_{i}$ in a patch-wise manner.

\begin{figure}[t]
\begin{center}
   \includegraphics[width=1.0\linewidth]{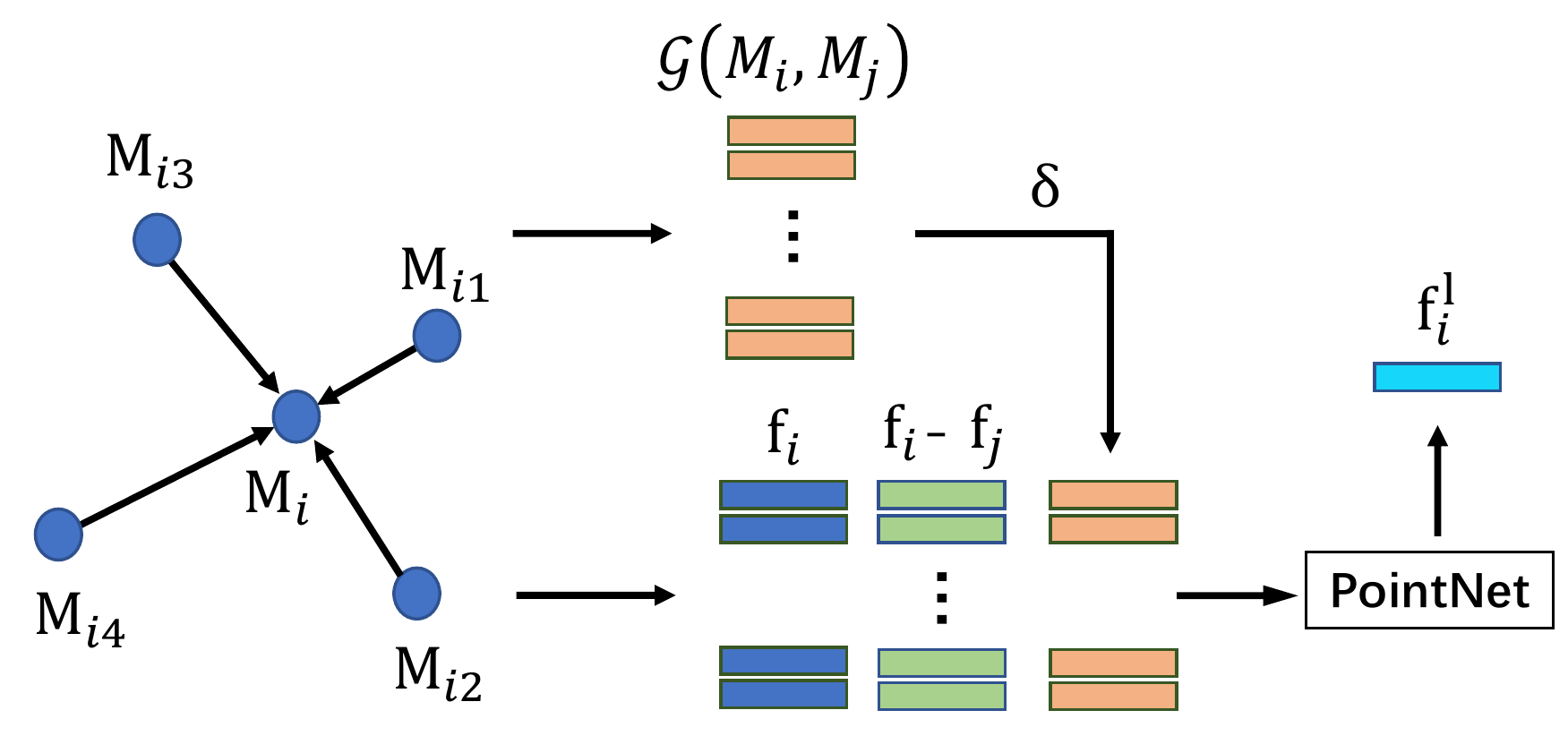}
\end{center}
   \caption{Illustration of the geometric relation embedded intra-scale edge convolution.}
\label{fig:intra-scale}
\end{figure}
\subsubsection{Intra-Scale Relative Pose Aware Feature Learning.}
To capture the geometry of patches within a larger receptive field, we apply the edge convolution~\cite{DGCNN} to further encode $\mathbf{f}_{i}$ to refine rotation-invariant features.
However, as mentioned before, the disentangled rotation-invariant shape feature does not contain pose information, performing convolution only on shape content features will lead to geometric information loss.
 As shown in Fig.~\ref{fig:intra-scale}, we address this issue by proposing a geometric relation encoding operation with a linear mapping function $\delta(\cdot)$ and embed the encoding into convolution, aggregating a local geometry feature $\mathbf{f}^{\ell}_{i}$ as follows:
\begin{equation} \label{orthloss}
\begin{aligned}
\mathbf{f}_{i}^{\ell} = \mathcal{A}_{j: j \in \mathcal{N}(i)}\Big(\operatorname{MLP}\big([\mathbf{f}_j, \mathbf{f}_{j}-\mathbf{f}_{i}, \delta(\mathcal{G}(\mathbf{M}_{i}, \mathbf{M}_{j}))]
\big)\Big),
\end{aligned}
\end{equation}
where $\mathcal{A}(\cdot)$ is the max-pooling.
In this way, we explicitly embed the relative pose information between different patches into the rotation invariance learning process, leading to a more distinct feature representation.

\subsubsection{Global-Scale Feature Disentangling.}
For global scale feature disentangling, we downsample $\mathbf{P}$ into a new point set $\mathbf{G} \in \mathbb{R}^{N_{g} \times 3}$ with $N_g$ points.
We further use local reference points $\mathbf{q}_{i} \in \mathbf{Q}$ to generate $N_{\ell}$ corresponding global patches.
Each global patch $\mathbf{M}_{i}^{g}$ will contain all points of $\mathbf{G}$ and the same translation operation in Eq.~\ref{translate} will be implemented to center the reference point and generate a reference frame.
In such a way, we integrate global context information and the corresponding local patch positional information 
into each global patch. 
Then we disentangle $\mathbf{M}_{i}^{g}$ into global-scale rotation-invariant features $\mathbf{f}^{g}_{i}$ and orientation matrices $\mathbf{O}^{g}_{i}$, which includes the shape information of global patches and the positional information of reference points. This information can be used in enhancing the feature distinction of rotation-invariant representations.
Besides, it is worth mentioning that there is no need to implement intra-scale learning for global scale patches, since the global patches cover all points, preserving the global context with the input shape.

\subsubsection{Inter-Scale Rotation-Invariant Learning.}
As local and global branches utilise the same downsampled point set $\mathbf{Q}$ as the reference point set, we discuss that local-scale and global-scale features $\mathbf{f}_{i}^{\ell}$ and $\mathbf{f}_{i}^{g}$ can be fused to combine local geometry and global context information.
To this end, we adopt geometric relations between $\mathbf{M}_{i}^{\ell}$ and $\mathbf{M}_{i}^{g}$ along with $\mathbf{f}_{i}^{\ell}$ and $\mathbf{f}_{i}^{g}$ as module inputs and directly output a relative pose-aware fused feature:

\begin{equation} \label{eq:fuse_feat}
\mathbf{f}_{i}^{out} = \mathcal{A}_{i: i \in \mathbf{Q}}\Big(\operatorname{MLP}\big([\mathbf{f}_{i}^{\ell}, \mathbf{f}_{i}^{g}, \delta(\mathcal{G}(\mathbf{M}_{i}^{\ell}, \mathbf{M}_{i}^{g}))]
\big)\Big).
\end{equation}
When calculating the geometric relation $\mathcal{G}(\mathbf{M}_{i}^{\ell}, \mathbf{M}_{i}^{g})$, we define the origin $\vec{\mathbf{0}}$ as reference points of global patches, otherwise reference points of $\mathbf{M}_{i}^{\ell}$ and $\mathbf{M}_{i}^{g}$ will be the same.

\subsection{Rotation-Invariant Pose-Aware Feature Propagation Module} \label{sec:3.3} 
\begin{figure}[t]
\begin{center}
   \includegraphics[width=1.0\linewidth]{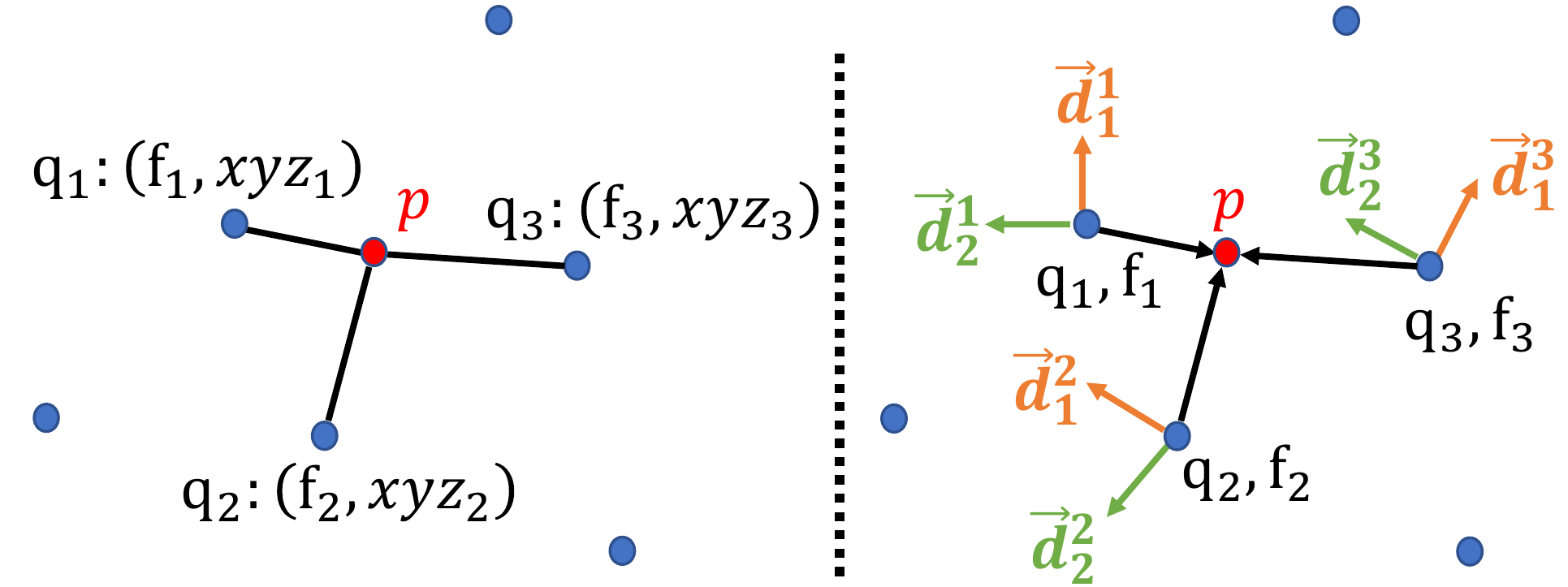}
\end{center}
   \caption{Left: typical feature propagation, the distance based interpolation. Right: relative pose-aware feature propagation process. The third direction $\vec{\mathbf{d}}_{3}$ of each patch is not shown for simplicity.}
\label{fig:long}
\label{fig:fp}
\end{figure}
Unlike the typical method proposed in PointNet++~\cite{PointNet++} (see Fig.~\ref{fig:fp} (left)), which propagates features from subsampled points to the original points based on point distances and absolute point positions over $k$ nearest neighbors, we propose a pose-aware feature propagation module, which dismisses distance-based interpolation and utilises relative geometric relations for pose information embedding.
As shown in Fig.~\ref{fig:fp} (right), given a point $\mathbf{q}_{j} \in \mathbf{Q}$ with feature $\mathbf{f}_{j}^{q}$ and a point $\mathbf{p}_{i} \in \mathbf{P}$ with feature $\mathbf{f}_{i}^{p}$, we follow Section~\ref{sec:3.2} to embed the feature propagation module with relative geometric relations between $\mathbf{q}_{j}$ and $\mathbf{p}_{i}$.
Specifically, a direction vector $\vec{\mathbf{u}}_{ij: j \in \mathcal{N}_{\mathbf{Q}}(i)} = \mathbf{q}_{j} - \mathbf{p}_{i}$ is defined, pointing from $\mathbf{p}_{i}$ to $\mathbf{q}_{j}$, where $\mathcal{N}_{\mathbf{Q}}(i)$ is the neighborhood of $\mathbf{p}_{i}$ in $\mathbf{Q}$.
As $\mathbf{q}_{j}$ is associated with a rotation-equivariant orientation matrix $\mathbf{O}_{j}$, we thus define a geometric relation between $\mathbf{p}_{i}$ and its neighbors $\mathbf{q}_{j}$ in the same way as Eq.~\ref{eq:geo_relation}:
\begin{equation} \label{eq:point_relation}
\mathcal{G}(\mathbf{p}_{i}, \mathbf{q}_{j}) = \left[\operatorname{dist}(\mathbf{p}_{i}, \mathbf{q}_{j}), \operatorname{cossim}(\mathbf{O}_{j}, [\vec{\mathbf{u}}_{ij}]_{j \in \mathcal{N}_{\mathbf{Q}}(i)})\right].
\end{equation}
The geometric relation $\mathcal{G}(\mathbf{p}_{i}, \mathbf{q}_{j})$ is then encoded by $\delta(\cdot)$ and concatenated with neighboring point features $\mathbf{f}^{q}_{j}$ and skip-linked point features $\mathbf{f}^{p}_{i}$ from the original points.
The fused features are passed through learnable MLPs and further aggregated by summation to update the original feature $\mathbf{f}^{p}_{i}$.
The whole process can be formulated as follows:
\begin{equation} \label{eq:fp}
\begin{aligned}
\mathbf{f}_{i} = \sum_{j=1}^{k} \operatorname{MLP}\left([\mathbf{f}^{p}_{i}, \mathbf{f}^{q}_{j}, \delta(\mathcal{G}(\mathbf{p}_{i}, \mathbf{q}_{j}))]\right).
\end{aligned}
\end{equation}
We can see from Eq.~\ref{eq:fp} that, unlike PointNet++, we ignore the rotation-sensitive global $xyz$ positions of point $\mathbf{p}_{i}$ during the propagation process, which ensures our module to be rotation-invariant.
Besides, the addition introduction of the neighboring point feature $\mathbf{f}^{q}_{j}$ further improves the feature representations.

\section{Experiments}
We evaluate our model with 3D point cloud classification and segmentation tasks, analyse the rotation robustness, compare the performance with other representative methods, and visualise the experimental results.
Furthermore, we analyse the efficiency of the proposed modules and the complexity of our model.
We follow the evaluation protocols of~\cite{Spherical}: we randomly rotate training and testing objects around z-axis (z/z), rotate training objects around z-axis while implement arbitrary rotations on testing objects (z/SO3), apply arbitrary rotations to both training and testing data (SO3/SO3).

For classification, we set $N_{\ell}$ to 256 and $N_g$ to 32.
When utilising $k$-NN for generating local-scale patches and searching neighbours for intra-learning, we assign the number of local-scale patches $k_{\ell}$ to 64 and the number of neighbours for intra-scale learning $k_{intra}$ to 32.
Settings for segmentation are same, except that $N_g$ and $k_{intra}$ are changed to 64 and 16, respectively.


\begin{table}  
\centering
\large
\resizebox{\linewidth}{!}{%
\begin{tabular}{l|cccc}
\hline
\bf{Rotation-Sensitive}  & input &  z/z    &z/SO3&SO3/SO3 \\ \hline
PointNet~\cite{PointNet}      & pc    & 85.9   & 19.6    &74.7  \\ 
PointNet++~\cite{PointNet++}    & pc  & 89.3   &  28.6    &85.0  \\ 
PointNet++~\cite{PointNet++}    & pc+n  & 91.8   & 18.4    &77.4  \\ 
DGCNN~\cite{DGCNN}   & pc  & \bf{92.2}   & 20.6    &81.1  \\ \hline
\bf{Rotation-Robust} & input &  z/z    &z/SO3&SO3/SO3 \\ \hline
Spherical CNN~\cite{Spherical}& Voxel  & 88.9   & 76.9    &86.9  \\ 
SFCNN~\cite{SFCNN}   & pc   & 91.4   & 84.8    &90.1  \\ 
RI-Conv~\cite{RIConv}   & pc & 86.5   & 86.4    &86.4  \\ 
ClusterNet~\cite{ClusterNet} & pc & 87.1   & 87.1    &87.1  \\ 
RI-GCN~\cite{RIgcn}& pc & 89.5   & 89.5    &89.5  \\ 
GCANet~\cite{Globalcontext} & pc & 89.0 & 89.1 &89.2\\
RIF~\cite{RIF}      & pc & 89.4   & 89.4    &89.3  \\ 
SGMNet~\cite{SGMNet}  & pc & 90.0   & 90.0    &90.0  \\ 
TFN~\cite{TFN2}   & pc    & 87.6   & 87.6    &87.6  \\ 
\citet{Poseselect}  & pc  & 90.2 & 90.2 &90.2 \\
VN-DGCNN~\cite{VN}      & pc    & 89.5   & 89.5    &90.2  \\ 
OrientedMP~\cite{messagepassing}    & pc    & 88.4   & 88.4    &88.9 \\
ELGANet~\cite{ELGANet}    & pc    & 90.3   & 90.3    &90.3 \\ \hline
\textbf{PaRot}   & pc     & 90.9 & \textbf{91.0}  &\textbf{90.8}  \\ \hline
\end{tabular}%
}

\caption{Classification performance on ModelNet40. ``pc'' and ``n'' denote the input datatype of raw 3D coordinates and normal, respectively.} \label{table:modelnet}
\end{table}

\subsection{Shape Classification}\label{sec:classification}
We test the classification ability of our model on the synthetic dataset ModelNet40~\cite{ModleNet40} and the real-world dataset ScanObjectNN~\cite{Objectnn}.
ModelNet40 is the most commonly used dataset in point cloud analysis, which contains 12,311 pre-aligned point cloud shapes sampled from 40 categories of CAD models. 
In the official version, the dataset is split into 9,843 training samples and 2,468 testing samples. 
ScanObjectNN contains 15000 incomplete objects scanned from 2,902 real-world objects.
To better explore the robustness to noise of scanned samples, we use the $OBJ\_BG$ subset which contains background noise for evaluation.
We sample 1024 points from each sample as our model inputs.

In our training procedure, we introduce random rotations to each patch in the disentanglement module.
During testing stage, no patch-wise rotation is applied, since disentangled features of our final model are rotation-robust, the performance will not be effected by patch-wise rotations.


We compare our model with representative models in terms of classification accuracy reported in Tables~\ref{table:modelnet} and \ref{table:objectnn} for ModelNet40 and ScanObjectNN respectively. It is shown that our method achieves promising results in all three settings with a high rotation-robustness as the absolute accuracy difference between z/z, z/SO3 and SO3/SO3 is not greater than $0.5\%$.

\begin{table}
\centering
\resizebox{\linewidth}{!}{%
\begin{tabular}{l|ccc}
\hline
Method       & z/SO3 & SO3/SO3 & $\Delta$    \\ \hline
PointNet~\cite{PointNet}     & 16.7  & 54.7    & 38.0   \\
PointNet++~\cite{PointNet++}   & 15.0    & 47.4    & 32.4 \\
DGCNN~\cite{DGCNN}       & 17.7  & 71.8    & 54.1 \\ \hline
RI-Conv~\cite{RIConv}     & 78.4  & 78.1    & 0.3  \\
RI-GCN~\cite{RIgcn}    & 80.5 & 80.6 & 0.1\\
RIF~\cite{RIF} & 79.8  & 79.9    & 0.1  \\
Li et al.$^{\star}$~(\citeyear{Poseselect})  &  79.3 & 79.6 & 0.3\\
VN-DGCNN$^{\star}$~\cite{VN} & 77.8 & 76.0 & 1.8\\ 
OrientedMP$^{\star}$~\cite{messagepassing} & 76.7 & 77.2 & 0.6 \\ \hline
\bf{PaRot}    & \textbf{82.1}  & \textbf{82.6}    & 0.5     \\ \hline
\end{tabular}%
}
\caption{Classification accuracy on ScanObjectNN $OBJ\_BG$ dataset under z/SO3 and SO3/SO3. $\Delta$ denotes the absolute difference between z/SO3 and SO3/SO3. $\star$ indicates our reproduced results based on official implementations.}
\label{table:objectnn}
\end{table}

\begin{table*}[t]
\centering
\LARGE
\resizebox{\linewidth}{!}{%

\begin{tabular}{l|c|cccccccccccccccc}

\hline
Method & C.mIoU & aero & bag & cap & car & chair & earph. & guitar & knife & lamp & laptop & motor & mug & pistol & rocket & skate & table \\ \hline
PointNet~\cite{PointNet}   & 37.8 & 40.4 & 48.1 & 46.3 & 24.5 & 45.1 & 39.4 & 29.2 & 42.6 & 52.7 & 36.7 & 21.2 & 55.0 & 29.7 & 26.6 & 32.1 & 35.8 \\
PointNet++~\cite{PointNet++}  & 48.3 & 51.3 & 66.0 & 50.8 & 25.2 & 66.7 & 27.7 & 29.7 & 65.6 & 59.7 & 70.1 & 17.2 & 67.3 & 49.9 & 23.4 & 43.8 & 57.6 \\
DGCNN~\cite{DGCNN}       & 37.4 & 37.0 & 50.2 & 38.5 & 24.1 & 43.9 & 32.3 & 23.7 & 48.6 & 54.8 & 28.7 & 17.8 & 74.4 & 25.2 & 24.1 & 43.1 & 32.3 \\ \hline
RI-Conv~\cite{RIConv}     & 75.3 & 80.6 & 80.0 & 70.8 & 68.8 & 86.8 & 70.3 & 87.3 & 84.7 & 77.8 & 80.6 & 57.4 & 91.2 & 71.5 & 52.3 & 66.5 & 78.4 \\
GCANet~\cite{Globalcontext}      & 77.2 & 80.9 & \textbf{82.6} & 81.0 & 70.2 & 88.4 & 70.6 & 87.1 & \bf{87.2} & \bf{81.8} & 78.9 & 58.7 & 91.0 & 77.9 & 52.3 & 66.8 & 80.3 \\
TFN~\cite{TFN2}         & 76.7 & 80.9 & 75.2 & 81.9 & 73.8 & 89.0 & 61.0 & 90.8 & 83.0 & 76.9 & 80.2 & 58.5 & 92.8 & 76.3 & 54.0 & \bf{74.5} & 79.1 \\
\citet{Poseselect} & 74.1 & 81.9 & 58.2 & 77.0 & 71.8 & \textbf{89.6} & 64.2 & 89.1 & 85.9 & 80.7 & \bf{84.7} & 46.8 & 89.1 & 73.2 & 45.6 & 66.5 & \bf{81.0} \\ 
VN-DGCNN$^{\star}$~\cite{VN}    &75.3 &81.1 &74.8 & 72.9 &73.8 &87.8 &55.9 &\textbf{91.4} &83.8 &80.2 &84.4 &44.5 &92.8 &74.6 &\textbf{57.2} &70.2 &78.9 \\ 
\hline
\bf{PaRot}  & \textbf{79.2} & \bf{82.7} & 79.2 & \textbf{82.3} & \textbf{75.3} & 89.4 & \textbf{73.9} & 91.1 & 85.6 & 81.0 & 79.5 & \bf{65.3} & \bf{93.9} & \bf{79.2} & 55.0 & 72.4 & 79.5 \\ \hline
\end{tabular}%
}
\caption{Segmentation per class results and averaged class mIoU on ShapeNetPart dataset under z/SO3, where C.mIoU stands for averaged mIoU of 16 classes.}
\label{table:shapenet}
\end{table*}

\subsection{Part Segmentation}\label{sec:segmentation}
\begin{figure}[t]
\begin{center}
   \includegraphics[width=\linewidth]{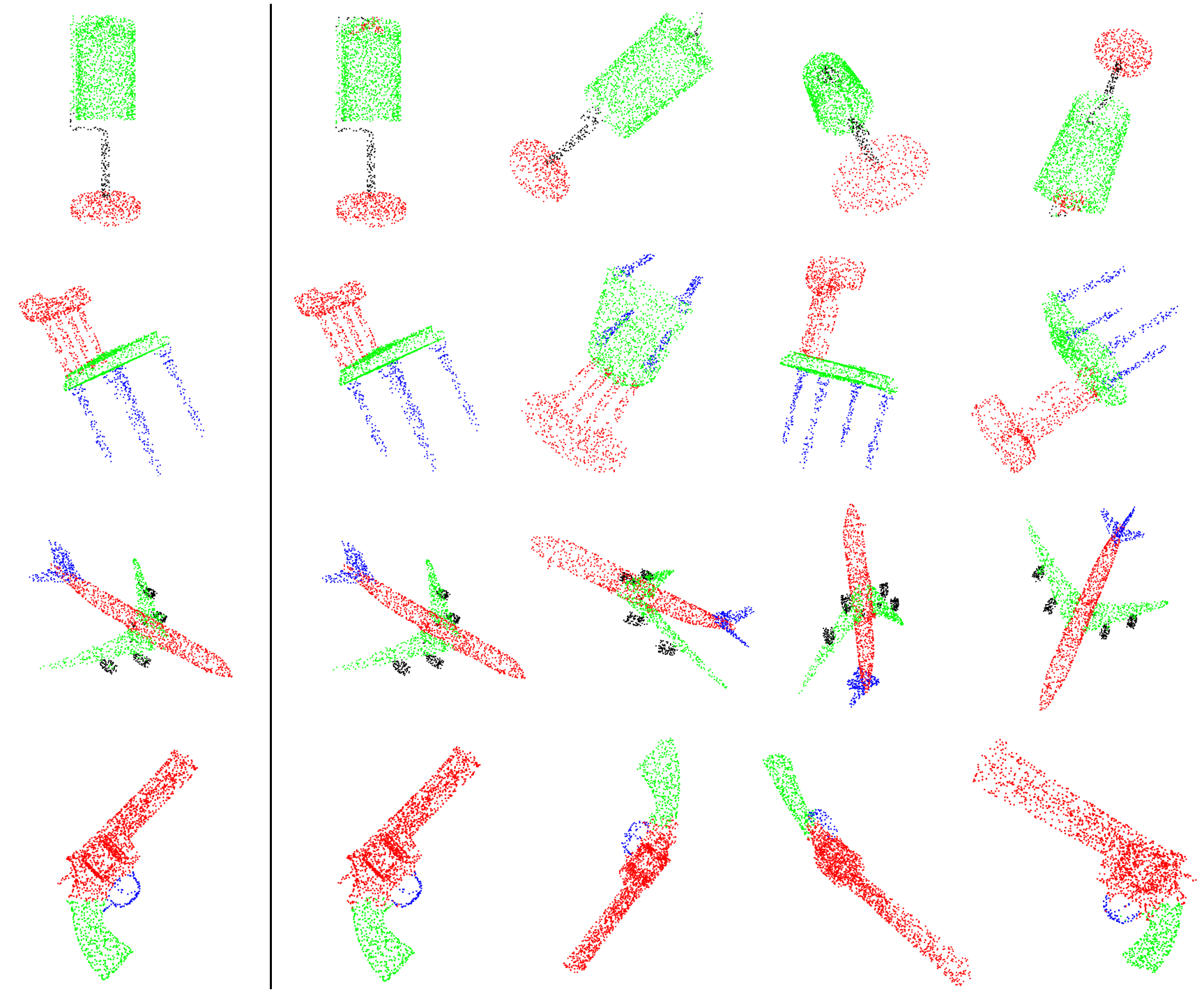}
\end{center}
\caption{Visualisation of ShapeNet part segmentation result. 
Left most column is ground truth of a selected orientation and the second left most is the segmentation result of the model with the select orientation. 
The rest of columns are results of random rotated samples.}
\label{fig:partseg}
\end{figure}

For shape part segmentation task, we validate our model on the ShapeNetPart dataset~\cite{ShapeNet}, which contains 16,880 synthetic samples with 14,006 training and 2,874 testing data.
The dataset includes 16 object categories and each category is annotated with 2 to 6 parts result in totally 50 part annotation labels.
2048 points are sampled as model input. The per-class mean intersection of union (mIoU) and the averaged mIoU of 16 classes under z/SO3 are reported and compared with other approaches in Table~\ref{table:shapenet}.
In addition, we visualise the segmentation results of different objects under different orientation in Fig.~\ref{fig:partseg}.
It is obvious that our model is robust in segmenting point clouds under arbitrary rotations.

\subsection{Method Analysis}\label{sec:methodanalysis}

\begin{table}[t]
\centering
\small
\begin{tabular}{l|ccccc|c}
\hline
\multirow{2}{*}{Model}  &
\multirow{2}{*}{Local} &
\multirow{2}{*}{Intra-} &
\multicolumn{1}{c|}{\multirow{2}{*}{Inter-}} &
\multicolumn{2}{c|}{Pose Rest.} &
\multirow{2}{*}{z/SO3} \\ \cline{5-6}
&  &  & \multicolumn{1}{c|}{} & orien. & position &  \\ \hline
\multicolumn{1}{c|}{A} & \checkmark  &             &            &  &               & 84.3  \\
\multicolumn{1}{c|}{B} & \checkmark  & \checkmark  &            &  &               & 87.4  \\
\multicolumn{1}{c|}{C} & \checkmark  &             & \checkmark &  &               & 70.9  \\
\multicolumn{1}{c|}{D} & \checkmark  & \checkmark  & \checkmark &  &               & 58.6  \\
\multicolumn{1}{c|}{E} & \checkmark  &             & \checkmark & \checkmark &  \checkmark   & 89.6  \\
\multicolumn{1}{c|}{F} & \checkmark  & \checkmark  &            & \checkmark &  \checkmark   & 89.5  \\ \cline{1-7}
\multicolumn{1}{c|}{G} & \checkmark  & \checkmark  & \checkmark & \checkmark &               & 90.4  \\
\multicolumn{1}{c|}{H} & \checkmark  & \checkmark  & \checkmark &            &  \checkmark   & 90.5  \\ 
\multicolumn{1}{c|}{I} & \checkmark  & \checkmark  & \checkmark & \checkmark &  \checkmark   & 91.0  \\ \hline
\end{tabular}
\caption{Component study results on ModelNet40 under z/SO3.}
\label{table:component}
\end{table}

\subsubsection{Component Study.}
We implement an ablation study to explore the effectiveness of each module in the proposed model and report results in Table~\ref{table:component}. 
We also split the geometric representation $\mathcal{G}(\mathbf{M}_m, \mathbf{M}_n)$ into two parts: $\operatorname{cossim}(\mathbf{O}_{m}, \mathbf{O}_{n})$ and $[\operatorname{dist}(\mathbf{p}_m, \mathbf{p}_n), \operatorname{cossim}(\mathbf{O}_{m}, [\vec{\mathbf{u}}_{mn}]), \operatorname{cossim}(\mathbf{O}_{n}, [\vec{\mathbf{u}}_{mn}])]$ to study the feature used for pose restoration. Here, $\operatorname{cossim}(\mathbf{O}_{m}, \mathbf{O}_{n})$ only contains relative orientation information and lacks positional information, while $[\operatorname{dist}(\mathbf{p}_m, \mathbf{p}_n), \operatorname{cossim}(\mathbf{O}_{m}, [\vec{\mathbf{u}}_{mn}]), \operatorname{cossim}(\mathbf{O}_{n}, [\vec{\mathbf{u}}_{mn}])]$ can detect relative position but has ambiguities for rotations around the vector $\vec{\mathbf{u}}_{mn}$.

Model A is our baseline that only extracts local-scale rotation-invariant shape content feature for classification.
When we employ intra-scale learning module (A$\rightarrow$B) without relative pose restoration operation, our model will suffer from pose information loss problem, however, it could aggregate features from a larger scale of patches and slightly improve the accuracy. This model has similar principle and performance of RI-Conv and ClusterNet.
The inter-scale learning module is designed for global context exploiting.
However, when adding inter-scale learning module without employing geometric relation representation (A$\rightarrow$C), the model performance drops, as the network cannot discover the relationship between the patch-wise features and the global context.
When using two learning modules (B$\rightarrow$D), the drop of performance is larger than model A$\rightarrow$C,  since B is deeper than A and is more vulnerable to the noisy feature added at deep layers.
The pose restoration strategy can preserve pose information, guide the learning process, and consistently improve the performance (B$\rightarrow$F, C$\rightarrow$E, D$\rightarrow$I).

Besides, models G, H and I show that both of two parts of geometric representation can restore part of the pose information and improve the accuracy, while the full representation will restore more information and achieve the best performance.

\begin{figure*}[t]
\begin{center}
   \includegraphics[width=1.0\linewidth]{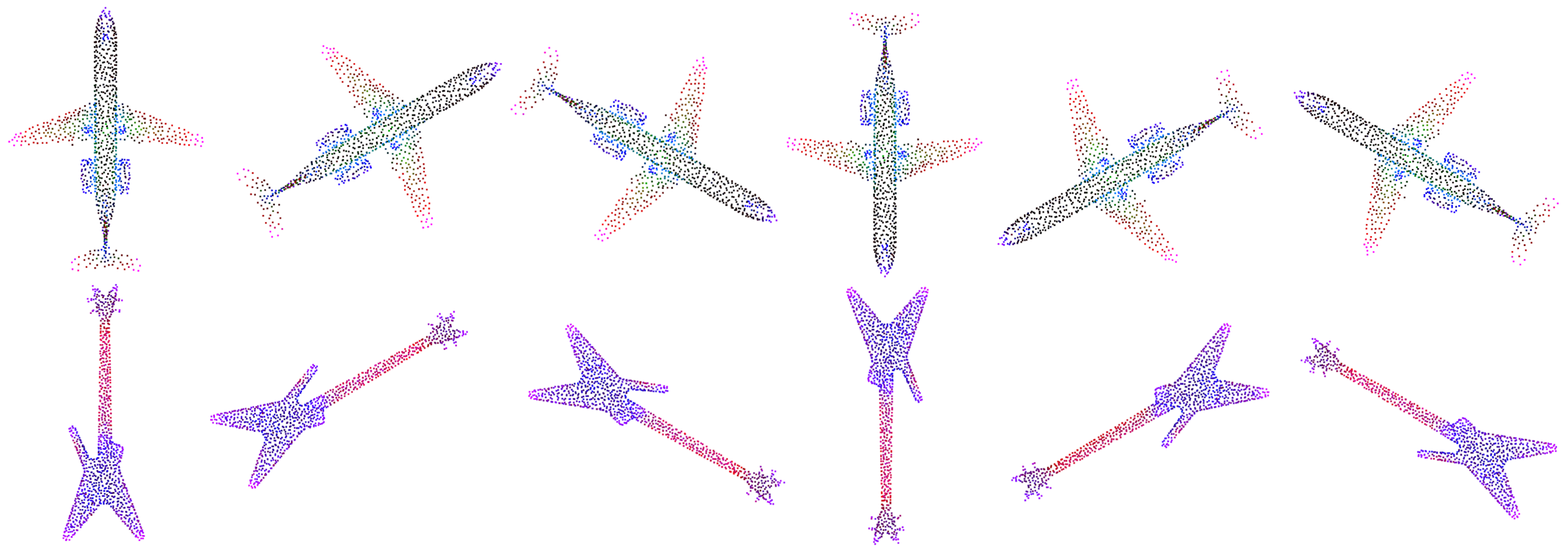}
\end{center}
   \caption{Visualisation of disentangled rotation-invariant features. From left to right, each sample is rotated $60^{\circ}$ around z-axis.}
\label{fig:inv_feat}
\end{figure*}

\subsubsection{Feature Propagation Analysis.}
To justify the proposed pose-aware geometric relation embedded feature propagation method, we compare the performance in terms of the mIoU over all instances (I.mIoU) and averaged mIoU of classes (C.mIoU) with the typical interpolation strategy introduced by PointNet++~\cite{PointNet++} in z/SO3 ShapeNet part segmentation task. 
The $xyz$ coordinate positions embedded in PointNet++ are removed, otherwise it will break the rotation invariance. 
As shown in Table~\ref{table:propagation}, our proposed propagation method outperforms interpolation strategy. 
Besides, the interpolation strategy weighted average the feature of nearest patches with respect to the inverse distance, patches with long distance contribute little to in generating the new feature and increasing the number of neighbours cannot improve the performance. In our method, the geometric relation between the target point and neighbour patches will be encoded individually, thus the performance can be further enhanced by increasing the number of neighbouring queries.

\begin{table}
\centering
\resizebox{\linewidth}{!}{%
\begin{tabular}{c|cc|cc}
\hline
\multicolumn{1}{l|}{}                  & \multicolumn{2}{l|}{Interpolation}                         & \multicolumn{2}{l}{Pose aware}                          \\ \hline
\multicolumn{1}{l|}{\# of neighbours}&\multicolumn{1}{l}{I.mIoU}&\multicolumn{1}{l|}{C.mIoU}&\multicolumn{1}{l}{I.mIoU}&\multicolumn{1}{l}{C.mIoU} \\ \hline
3  & 81.1  & 76.6  & 82.1          & 77.6          \\
5  & 80.9  & 76.7  & 82.5          & 78.3          \\
7  & 80.8  & 76.8  & 82.8          & 78.1          \\
9  & 80.8  & 76.7  & 83.0          & 79.1          \\
11 & 80.8  & 76.7  & 82.9          & \textbf{79.2} \\
13 & 80.6  & 76.4   & \textbf{83.2} & 79.0          \\ \hline
\end{tabular}%
}
\caption{Propagation analysis. Results of Instance mIoU and Class averaged mIoU of z/SO3 part segmentation on ShapeNetPart. The left most column determines the number of neighbour patches used for feature propagation.}
\label{table:propagation}
\end{table}

\subsubsection{Model Complexity.}
Benefiting from the efficient hierarchical structure and geometric relation encoding technique, our rotation-robust model achieves high performance with a small model size and low computational cost. We avoid the calculation of dynamic graphs~\cite{DGCNN} and balance the width and depth of the network during developing procedure. Moreover, siamese procedure is not necessary during testing, thus it could be removed to further reduce the computational complexity. As shown in Table~\ref{table:complexity}, the proposed model improves the accuracy by $1.5\%$ with only $55\%$ parameters and $45\%$ FLOPs compared to VN-DGCNN.

\begin{table}
\centering
\resizebox{\linewidth}{!}{%
\begin{tabular}{l|ccc}
\hline
Method       & Para. & FLOPs & Acc.   \\ \hline
PointNet++~\cite{PointNet++}   & 1.41M & 863M & 85.0 \\
DGCNN~\cite{DGCNN}        & 1.72M  & 2449M  & 81.1 \\ \hline
RI-GCN~\cite{RIgcn}        & 4.19M  &  1237M  & 89.5  \\
RIF~\cite{RIF}        & 2.36M  &  6535M  &  89.4  \\
\citet{Poseselect}        & 2.91M  & 3747M &  90.2  \\
VN-DGCNN~\cite{VN}      & 2.77M  & 3183M & 89.5  \\ \hline
PaRot-training         & 1.55M  &  2091M   & 91.0 \\
PaRot-testing       & 1.55M  &  1431M  &  91.0 \\ \hline
\end{tabular}%
}
\caption{Comparisons of model size, computational complexity, and z/SO3 accuracy on ModelNet40. The \textit{testing} version removes unnecessary auxiliary network modules.}
\label{table:complexity}
\end{table}
\subsubsection{Disentangled Rotation-Invariant Feature Visualisation.}
To examine the effectiveness of our disentanglement module in extracting consistent features for patches under arbitrary orientation, we visualise the feature responses of two different objects (i.e., aeroplane and guitar), which are rotated around the z-axis.
Specifically, three channels of the disentangled rotation-invariant features which are extracted from the local branch are selected and taken as the RGB channel for visualisation.
We generate 1024 local-scale patches with all points in $P$ as reference points and disentangle with saved best segmentation model.
The result in Fig.~\ref{fig:inv_feat} indicates that the features learned are various for patches with different shape content while invariant to rotations.
\section{Conclusion}
In this work, we propose PaRot which is a novel rotation-invariant learning model for 3D point cloud recognition.
Given a point cloud, we disentangle rotation-invariant shape content features and rotation-equivariant orientations for local-scale patches and global-scale patches by introducing pairs of rotations under a siamese training procedure. To restore the rotation-sensitive pose information while maintaining the rotation invariance of learning, we compute geometric relations with patch-wise orientation matrices to represent the relative pose between patches. The geometric relations are utilised to guide the intra-scale and inter-scale feature aggregation. Following the same idea of restoring pose information with geometric relations, we further design a rotation-invariant feature propagation method which improves the segmentation accuracy of our model. Extensive experiments demonstrate the effectiveness and efficiency of our model.

\bibliography{preprint}

\end{document}


\maketitle

In the supplementary material, we first introduce the details about the training procedure and the network structures of the PaRot architectures in Section 1.
We then discuss the effect of three loss functions proposed in the disentanglement modules in Section 2.
Section 3 illustrates the restored pose feature of the inter-scale learning.
Finally, we provide additional ablation study results of hyperparameter setting in Section 4 and experimental results related to segmentation in Section 5.

\section{Implementation Details}
\subsection{Experimental Setting}

The model is evaluated with PyTorch in Nvidia RTX3090. Settings about generating patches are introduced in the main paper.

Our total loss function $\mathcal{L}_{total}$ is defined as:
\begin{equation} \label{eq:total_loss}
\begin{aligned}
    \mathcal{L}_{total} = \mathcal{L}_{cls} + \alpha_\ell \mathcal{L}_{equi\_\ell} + \mathcal{L}_{orth\_\ell} + \beta_\ell \mathcal{L}_{inv\_\ell} \\ 
    \alpha_g\mathcal{L}_{equi\_g} + \mathcal{L}_{orth\_g} + \beta_g\mathcal{L}_{inv\_g},
\end{aligned}
\end{equation}
where $\mathcal{L}_{cls}$ is the cross-entropy classification loss, and the subscripts $\ell$ and $g$ denote the loss function belonging to local-scale and global-scale disentanglement modules respectively.
Moreover, $\alpha_{\ell} $, $\alpha_g$, $\beta_{\ell}$, and $\beta_g$ are the weighting parameters adjusting the contribution of different loss functions from local and global scales.
We set $\alpha_\ell$, $\alpha_g$, $\beta_\ell$, and $\beta_g$ to 0.2, 0.1, 0, and 0, respectively.
The reason why we set the invariant loss function ${L}_{inv}$ to 0 is discussed in Section~\ref{sec:loss}.

For classification, the input point clouds are randomly scaled in the range of [0.67, 1.5] for augmentation during training, and the training epoch is 250 with batch size of 32.
Adam optimizer is utilised and the learning rate is initialised to 1e-3, scheduled to 1e-5 with cosine annealing scheduler. The momentum and weight decay are set to 0.9 and 1e-6 respectively.
For segmentation, the experimental settings are the same as those of classification, except that $N_g$ is change to 64 and $k_{intra}$ to 16. 
We concatenate the one-hot class label vector to the last feature layer following the implementation of PointNet++~\cite{PointNet++}.

\begin{figure}
\centering
   \includegraphics[width=1.0\linewidth]{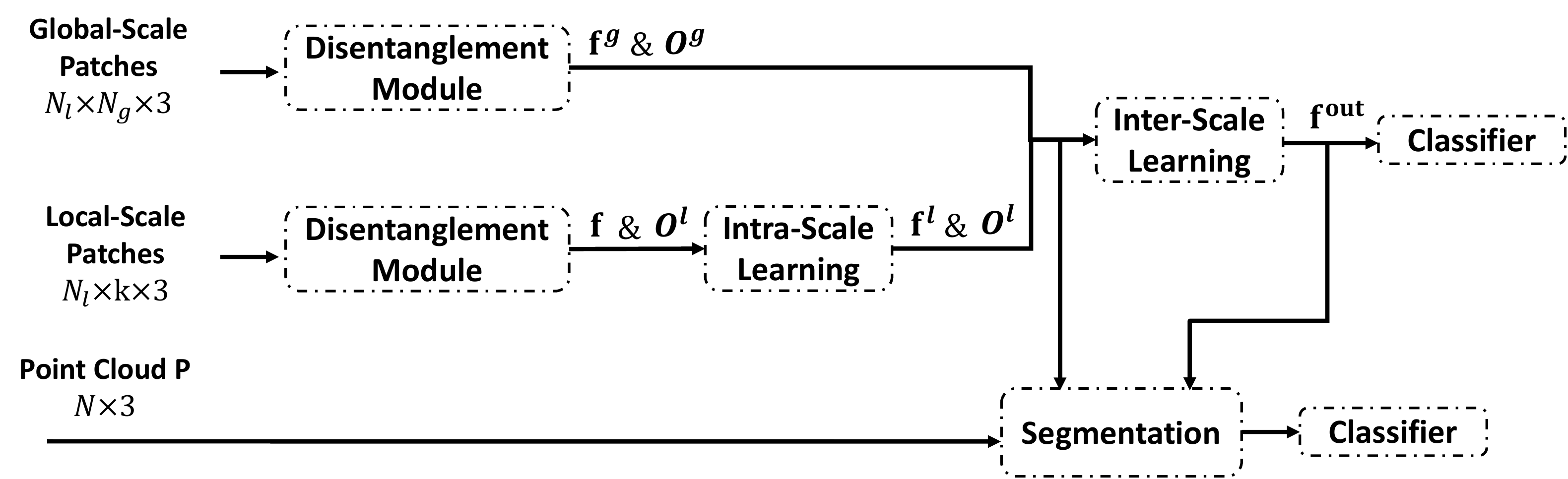}
   \caption{The overall structure of PaRot.}
   \label{fig:overall}
\end{figure}

\begin{figure}
\begin{center}
   \includegraphics[width=1.0\linewidth]{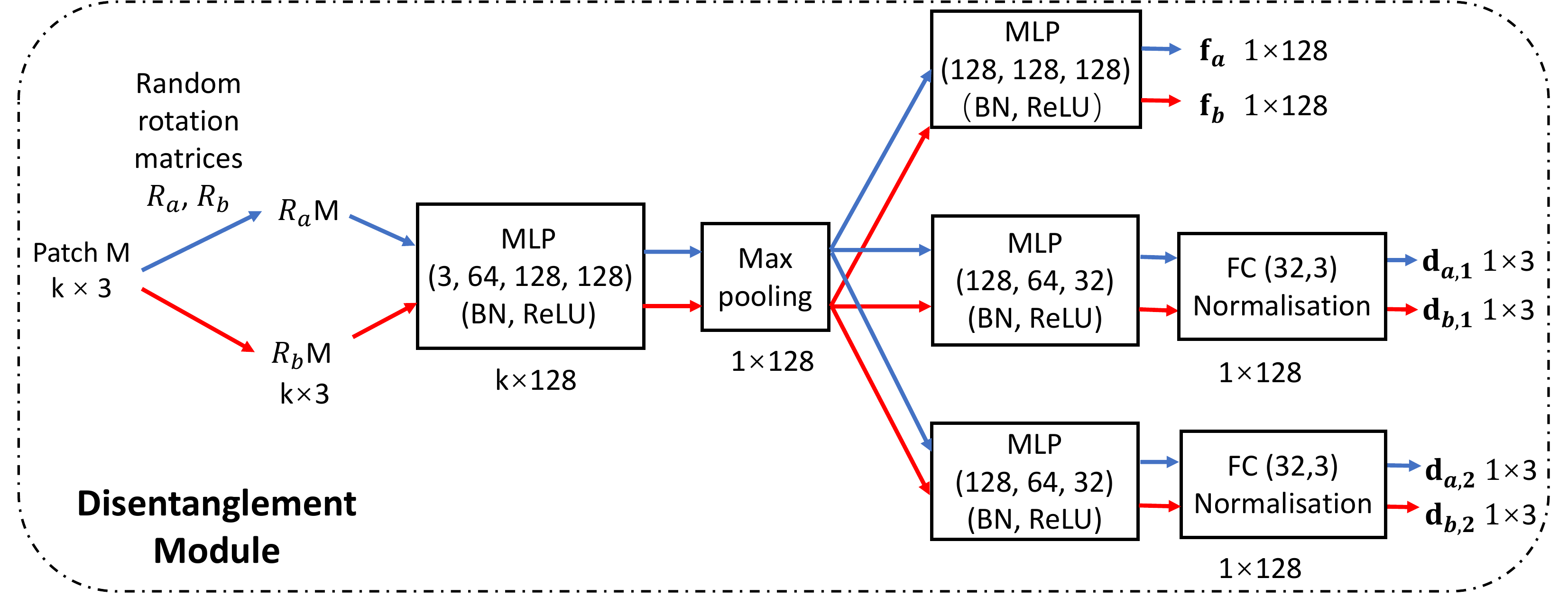}
\end{center}
   \caption{Detailed architecture of the proposed disentanglement module. Two disentanglement modules with similar architecture are assigned to process local-scale patches and global-scale patches independently. }
\label{fig:disentangle_module}
\end{figure}

\begin{figure}
\begin{center}
   \includegraphics[width=1.0\linewidth]{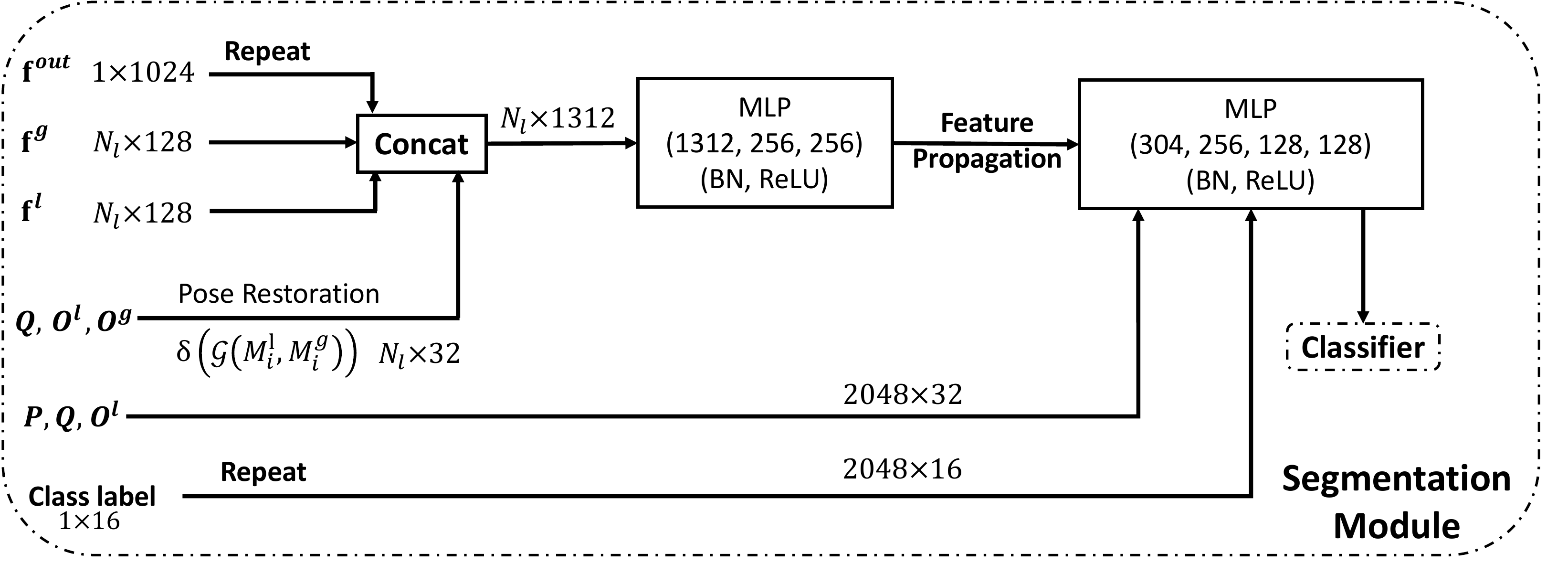}
\end{center}
   \caption{Detailed architecture of the PaRot segmentation module.}
\label{fig:segment_module}
\end{figure}

\subsection{Model Architecture}

The overall architecture of the PaRot model is illustrated in Fig.~\ref{fig:overall}. The details about disentanglement module and segmentation module are presented in Fig.~\ref{fig:disentangle_module} and Fig.~\ref{fig:segment_module} respectively.
The architectures of intra-scale learning module, inter-scale learning module and feature propagation module are relative simple and have been explained in the main work, therefore we only provide the information with written expressions.
It is worth mentioning that all geometric relation encoding functions, \textit{i.e.}, $\delta(\cdot)$, consist of a fully connected layer with output channel number of 32, a batch normalisation layer, and an ReLU.

In the intra-scale learning module, we implement EdgeConv~\cite{DGCNN} by performing $k$-nn search within Euclidean space, where we take as input the rotation-invariant features from two patches as well as the encoded pose feature.
The channel numbers of the MLP inside intra-scale learning module are 288, 128, 128, and 128.
The inter-scale learning module is responsible for both feature aggregation and channel raising, thus the channel of MLP is set to 288, 256, 512, 1024 with LeakyReLU (0.2) as the activation function.
The classifiers for classification and segmentation are borrowed from PointNet++~\cite{PointNet++}.

\section{Loss Function} \label{sec:loss}
The proposed disentanglement module contains three loss functions, constraining the learned features being either rotation-invariant or rotation-equivariant.
In this section, we implement an ablation study to investigate the effectiveness of three loss functions \textit{i.e.}, Eqs.~(1-3) of the main work on our model performance.



\begin{table}[b]
\center
\begin{tabular}{l|ccc|c}
\hline
Model & $\mathcal{L}_{inv}$ & $\mathcal{L}_{equi}$ & $\mathcal{L}_{orth}$ &  Acc. \\ \hline
\multicolumn{1}{l|}{A} & \checkmark  & \checkmark  & \checkmark   &  90.6 \\
\multicolumn{1}{l|}{B} &             & \checkmark  & \checkmark   &  \bf{91.0}\\ 
\multicolumn{1}{l|}{C} & \checkmark  &             &              &  90.6   \\
\multicolumn{1}{l|}{D} &             & \checkmark  &              &  90.3 \\
\multicolumn{1}{l|}{E} &             &             & \checkmark   &  88.6   \\
\multicolumn{1}{l|}{F} &             &             &              &  90.4 \\ \hline
\end{tabular}
\caption{Ablation study on loss functions in our disentanglement module. Results on ModelNet40 under z/SO3 are reported.}
\label{table:loss}
\end{table}

\begin{figure*}
\centering
   \includegraphics[width=1.0\linewidth]{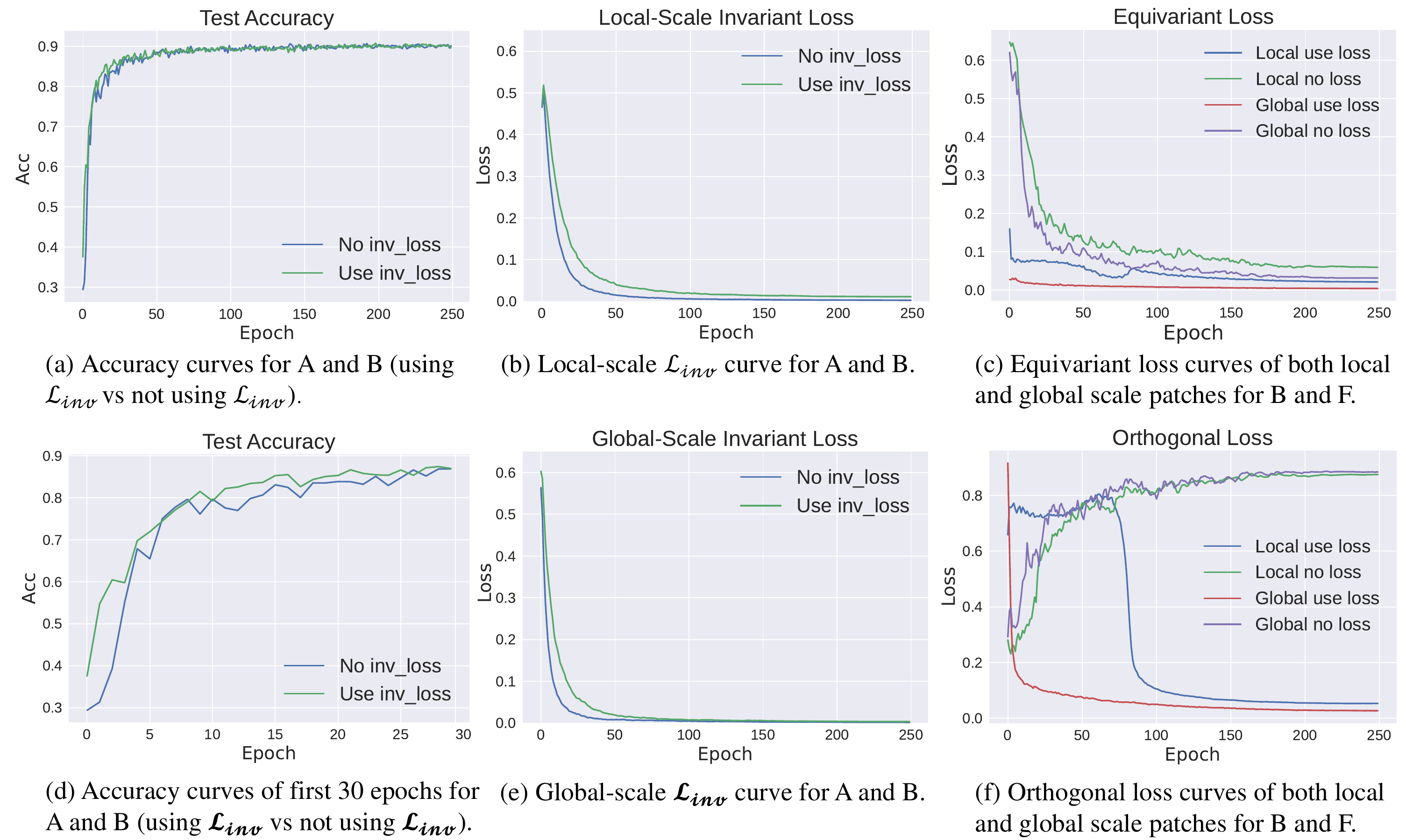}
\caption{Accuracy and loss curves for model A, B, and F in Table~\ref{table:loss}.}
\label{fig:loss}
\end{figure*}

As shown in Table~\ref{table:loss}, when no restrictions are applied, the classification accuracy (F) is 90.4\%, which is lower than our best model B with the accuracy of 91.0\%.
Fig.~\ref{fig:loss} (c) and (f) show that the combination of $\mathcal{L}_{equi}$ and $\mathcal{L}_{orth}$ speeds up the learning of rotation-equivariant vectors and sufficiently enforces two vectors to be perpendicular to each other, which improves the accuracy by 0.6\%.
Comparing models C-E with B, it is clear that the single application of the loss function cannot achieve the best result.
Moreover, we find that when applying all the loss functions (model A), the model performance drops compared to model B.
As shown in Fig.~\ref{fig:loss} (a), (b), (d) and (e), the $\mathcal{L}_{inv}$ of model A decreases faster than model B and has a better performance in first 30 epochs. However, when the number of epoch is sufficiently large, $\mathcal{L}_{inv}$ will hinder the learning of shape content feature and result in a drop of accuracy.




To further analyse the effect of implementing the combination of  $\mathcal{L}_{equi}$ and $\mathcal{L}_{orth}$, we visualise the equivariant loss curve and the orthogonal loss curve of B and F in Fig.~\ref{fig:loss} (c) and (f). 
When $\mathcal{L}_{equi}$ and $\mathcal{L}_{orth}$ are not used, the equivariant loss will still decrease slowly but the $\mathcal{L}_{orth}$ will keep increasing, which means the learned two direction vectors are parallel to each other and it will cause some ambiguity problems when restoring the pose information. 
In addition, it shows that learning orientation matrices for local-scale patches are more difficult than for global-scale patches.
 

\begin{figure*}
   \includegraphics[width=1.0\linewidth]{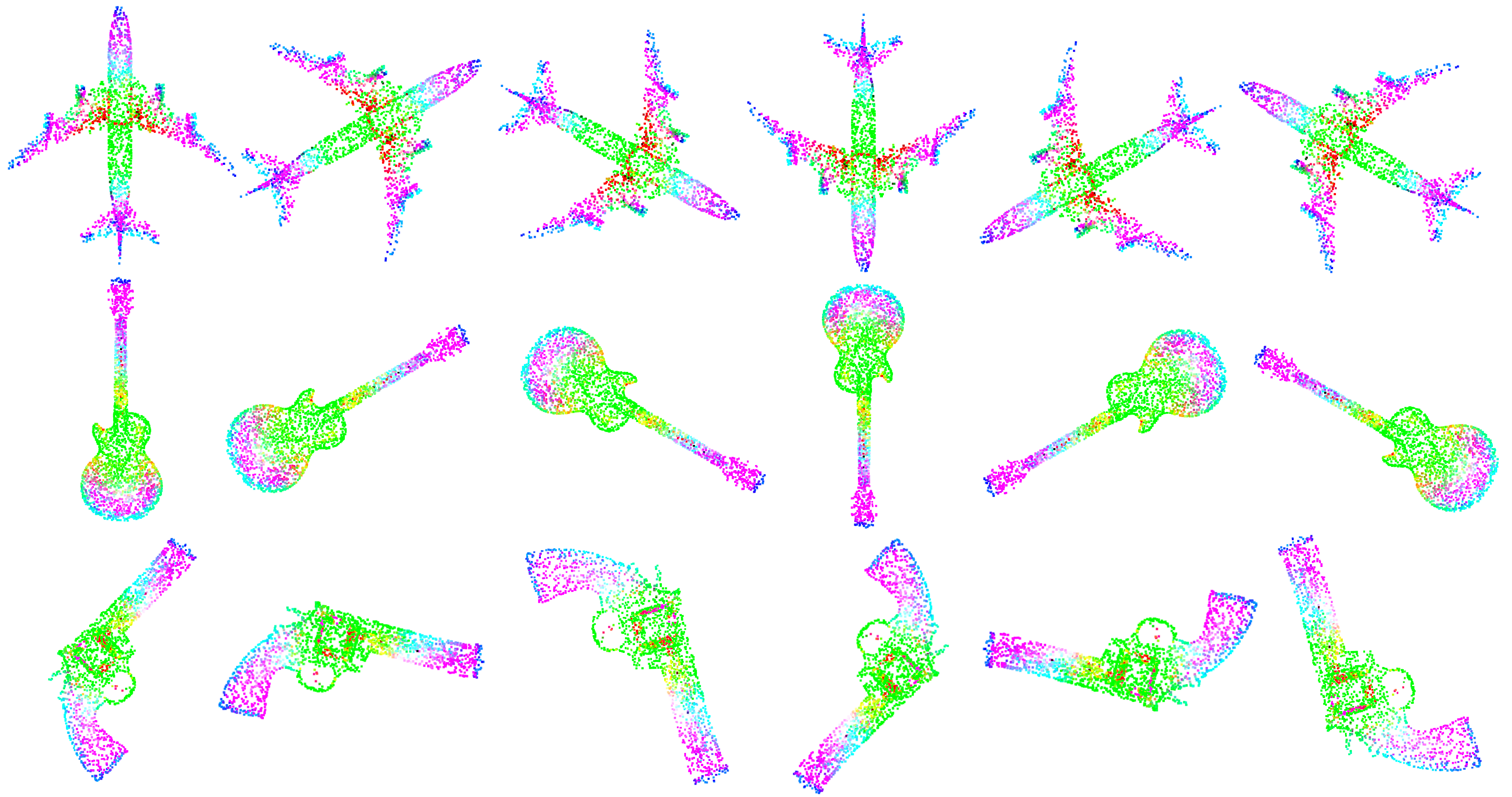}
   \caption{Visualisation of restored pose features in inter-scale learning. From left to right, each sample is rotated $60^{\circ}$ around z-axis.}
\label{fig:restore_pose_rotfig}
\end{figure*}
\section{Restored Pose Feature Visualisation}
To examine the rotation invariance and effectiveness of our restored pose information, we visualise the restored pose features of inter-scale learning module in ShapeNet part segmentation task.
We follow the same procedure of visualising disentangled feature in the main paper, selecting three channels as the RGB values and choose three objects from aeroplane, guitar, and pistol class rotating around z-axis for visualisation.
We set $N_{\ell}$ to 2048 to provide dense results with saved models.

As it has been discussed in the main paper, the pose restoration module for inter-scale learning aims to explore the relationship between the patch-wise features and the global context. The learned features need to be rotation-invariant and contain both relative positional information and patch-wise orientation information.
As illustrated in Fig.~\ref{fig:restore_pose_rotfig}, restored features are consistent under different orientations. Besides, areas close to centers of objects are generally painted with green, while farther areas are painted in pink. In addition, effected by the orientation of specific patches, some marginal areas are presented in blue and some complicated patches (i.e. the wing-fuselage connection joint and the wheel of pistol) are shown in red.



\section{Ablation Study}
\begin{table}
\small
\center
\begin{tabular}{ccc|ccc}
\hline
searching  & radius & $k_{\ell}$ & z/z & z/SO3 & FLOPs \\ \hline
ball query    & 0.2  & 64   & 90.3 &  90.4 &  1431M \\
ball query    & 0.3  & 64   & 90.3 & 90.5 & 1431M \\
knn     &      -     & 32   &  90.7 & 90.6 & 1220M \\
knn      &     -     & 64   & \bf{90.9} & \bf{91.0} & 1431M \\
knn       &    -     & 128  & 90.8 & 90.6 &  1852M \\ \hline
\end{tabular}
\caption{Ablation study on generation of local-scale patches. Results on ModelNet40 under z/z, z/SO3.}
\label{table:local}
\end{table}
\subsection{Generation of Local-scale Patches} \label{sec：localpatch}
There are two neighbor search methods investigated for generating local-scale patches: $k$-NN search and ball query.
In Table~\ref{table:local}, we examine both methods and report their corresponding results with different numbers of neighbors extracted, where we find that models utilising $k$-nn outperform models employing ball query method. This might because the ball query method strictly constrains the size of generated patches, and $k$-nn method is more flexible and would generate better patches from sparse and dense parts of point clouds.
When setting $N_{\ell}$ to 64, the $k$-nn based model can achieve the best performance, and the computational cost is also moderate.


\begin{table*}
\LARGE
\resizebox{\linewidth}{!}{
\begin{tabular}{l|c|cccccccccccccccc}
\hline
Method & C.mIoU & aero & bag & cap & car & chair & earph. & guitar & knife & lamp & laptop & motor & mug & pistol & rocket & skate & table \\ \hline
PointNet~\cite{PointNet}   & 74.4 &81.6& 68.7& 74.0& 70.3& 87.6& 68.5& 88.9& 80.0& 74.9& 83.6& 56.5& 77.6& 75.2& 53.9& 69.4& 79.9 \\
PointNet++~\cite{PointNet++}  & 76.7 &79.5& 71.6& \bf{87.7}& 70.7& 88.8& 64.9& 88.8& 78.1& 79.2& \bf{94.9}& 54.3& 92.0& 76.4& 50.3& 68.4& 81.0 \\
DGCNN~\cite{DGCNN}       &73.3& 77.7& 71.8& 77.7& 55.2& 87.3& 68.7& 88.7& 85.5& \bf{81.8} & 81.3& 36.2& 86.0& 77.3& 51.6& 65.3& 80.2 \\ \hline

RI-Conv\cite{RIConv}   & 75.3 & 80.6& 80.2& 70.7& 68.8& 86.8& 70.4& 87.2& 84.3& 78.0& 80.1& 57.3& 91.2& 71.3& 52.1& 66.6& 78.5\\
GCANet~\cite{Globalcontext}     & 77.3 & 81.2 & \bf{82.6} & 81.6 & 70.2 & 88.6 & 70.6 & 86.2 & \bf{86.6} &  81.6 & 79.6 & 58.9 & 90.8 & 76.8 & 53.2 & 67.2 & \bf{81.6} \\

TFN~\cite{TFN2}        & 78.4 & 80.3 & 77.3 & 82.6 & 74.7 & 88.8 & \bf{76.3} & 90.7 & 81.7 & 77.4 & 82.4 & \bf{60.7} & 93.2 & 79.4 & 54.3 & \bf{74.7} & 79.6 \\
\citet{Poseselect} & 74.1 & 81.9 & 58.2 & 77.0 & 71.8 & \bf{89.6} & 64.2 & 89.1 & 85.9 & 80.7 & 84.7 & 46.8 & 89.1 & 73.2 & 45.6 & 66.5 & 81.0 \\ 
VN-DGCNN$^{\star}$~\cite{VN}     &75.4 &81.0 &76.1 &76.0 &71.4 &88.1 &59.4 &91.3 &85.0 &80.4 &85.5 &44.7 &92.3 &74.5 &52.4 &68.7 &78.9 \\
\hline

PaRot  & \bf{79.5} & \bf{82.9} & 82.1 & 83.2 & \bf{75.7} & 89.4 & 76.1 & \bf{91.5} & 86.1 & 81.4 & 80.3 & 59.3 & \bf{94.3} & \bf{79.7} & \bf{57.0} & 73.3 & 79.2 \\ \hline
\end{tabular}}
\caption{Segmentation per class results and averaged class mIoU on ShapeNet Part dataset under SO3/SO3. $\star$ indicates our reproduced results based on official implementations.}
\label{table:segment_so3_so3}
\end{table*}

\begin{figure*}[t]
\begin{center}
   \includegraphics[width=\linewidth]{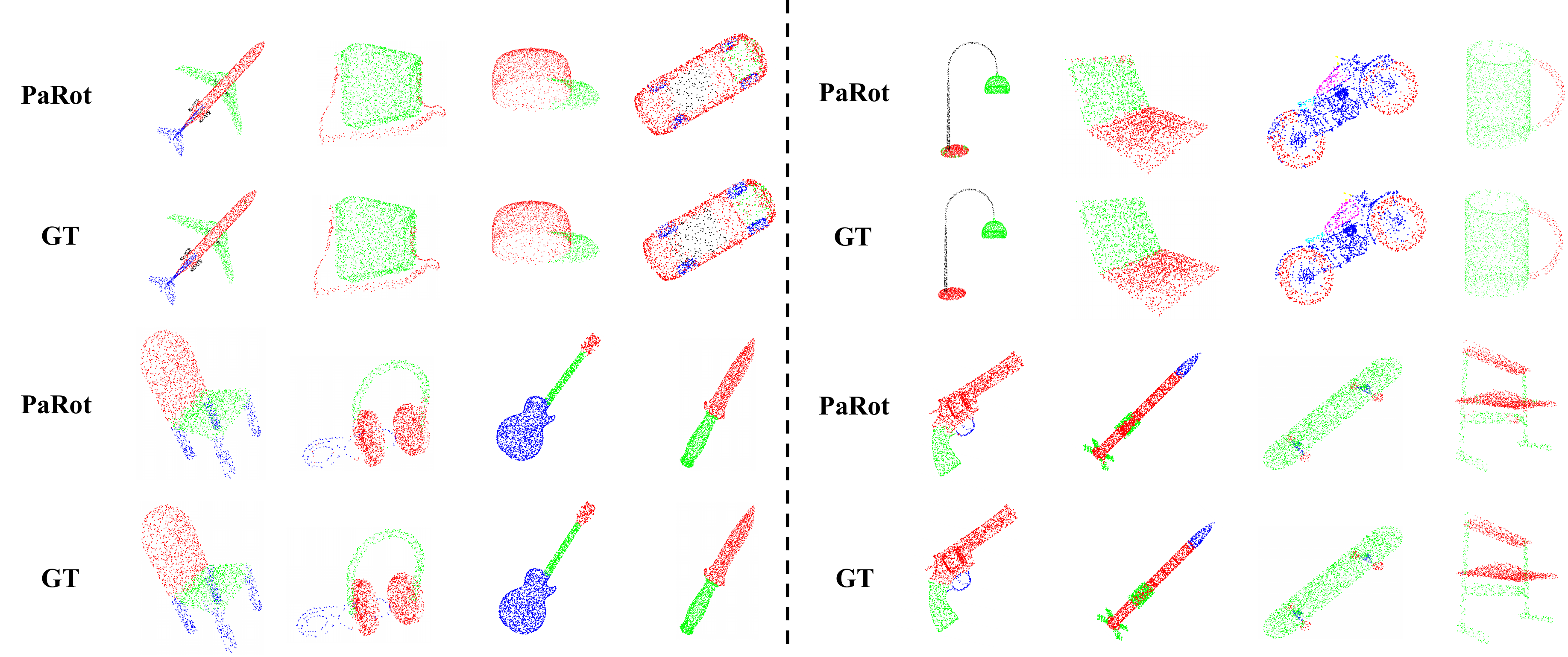}
\end{center}
   \caption{Comparisons between ground truth (GT) annotations and the outputs generated by PaRot for z/SO3 ShapeNet Part segmentation task.}
\label{fig:segmentation_GT}
\end{figure*}

\begin{table}
\centering
\resizebox{0.95\linewidth}{!}{
\begin{tabular}{ccc|ccc}
\hline
$N_{\ell}$ & $k_{intra}$ & $N_g$   & z/z & z/SO3 & FLOPs \\ \hline
64      & 32 & 32 & 90.1 & 90.0 &  358M \\
128     & 32 & 32 & 90.8 & 90.5 & 716M  \\
256     & 32 & 32 & \bf{90.9} & \bf{91.0} &  1431M \\
512     & 32 & 32 & 90.6 & 90.5 & 2861M  \\  \hline
256     & 8 & 32  &  90.3 & 90.3  &  995M \\ 
256     & 16 & 32  &  90.5 & 90.5  &  1140M \\ 
256     & 64 & 32  & 90.6 & 90.5 & 2012M \\ \hline
256     & 32 & 8  & 90.7 & 90.7 &  1273M \\
256     & 32 & 16  & 90.7 & \bf{91.0} &  1325M \\
256     & 32 & 64  & 90.6 & 90.5 & 1641M \\ \hline

\end{tabular}
}
\caption{Ablation studies on $N_{\ell}$,  $N_g$, and $k_{intra}$. Experiments are conducted on ModelNet40 under z/z, z/SO3.}
\label{table:ablation}
\end{table}

\subsection{Hyperparameter Selection}
We have investigated $k_{\ell}$ in Section~\ref{sec：localpatch} and there are three other hyperparameters, \textit{i.e.}, the number of patches to generate $N_{\ell}$, the number of points in global-scale patches $N_g$, and the numbers of neighbours to query in intra-learning $k_{intra}$. To analyse the impact of those three hyperparameters, we conduct more experiments on ModelNet40 and results are shown in Table~\ref{table:ablation}. 

For the number of patches to generate, if we set $N_l$ to be a small value, the generated patches will not be able to cover all the parts of point cloud and results in the reduction of accuracy. However, setting $N_l$ to a very large value will not only significantly increase the computational cost, but also reduce the receptive field of intra-scale learning and harm the performance.
The value of $k_{intra}$ also has a high influence to the computational cost, and we found that when $N_l=256$, setting $k_{intra}$ to 32 will achieve the best performance.
Ablation studies on the $N_g$ (number of points sampled for global scale patches) show that the proposed methods can restore efficient inter-scale pose information when only using 8 points for global patches and it can substantially reduce the computational cost. 
Besides, Table~\ref{table:ablation} also shows that we can further reduce the computational cost of PaRot by modifying hyperparameters while maintaining a high accuracy.




\section{Additional Segmentation Results}
We report the results of ShapeNet Part segmentation SO3/SO3 in terms of the per-category mIoU in Table~\ref{table:segment_so3_so3}. 
It is shown that typical rotation-sensitive models perform much better in SO3/SO3 than in z/SO3. By augmenting the training samples with rotations, typical models can outperform some rotation-robust methods in segmentation tasks, especially in class cap, lamp, and laptop.
However, the proposed PaRot method still outperforms these methods in terms of averaged class mIoU and achieves balance performance among all 16 classes.
We also visualise one sample from each object class with our trained z/SO3 model in Fig.~\ref{fig:segmentation_GT}. Although we can detect some segmentation errors when comparing the ground truths and predicted samples, PaRot provides accurate predictions in most classes.
\bibliography{supplementary}